\documentclass[letterpaper]{article} 
\usepackage{aaai25}  
\usepackage{times}  
\usepackage{helvet}  
\usepackage{courier}  
\usepackage[hyphens]{url}  
\usepackage{graphicx} 
\usepackage{natbib}  
\usepackage{caption} 
\usepackage{algorithm}
\usepackage{algorithmic}
\usepackage{multirow}
\usepackage{pifont}
\usepackage{subcaption}
\usepackage{newfloat}
\usepackage{listings}
\DeclareCaptionStyle{ruled}{labelfont=normalfont,labelsep=colon,strut=off} 
\floatstyle{ruled}
\newfloat{listing}{tb}{lst}{}
\floatname{listing}{Listing}
\usepackage{amsmath}

\usepackage[capitalize]{cleveref}
\crefname{section}{Sec.}{Secs.}
\Crefname{section}{Section}{Sections}
\crefname{subsection}{Sec.}{Secs.}
\Crefname{subsection}{Section}{Sections}
\Crefname{table}{Table}{Tables}
\crefname{table}{Tab.}{Tabs.}

\usepackage{booktabs}
\usepackage[table,xcdraw]{xcolor}
\usepackage{cleveref}
\usepackage{dsfont}

\title{PAT: Pruning-Aware Tuning for Large Language Models}
\author{
    Yijiang Liu\textsuperscript{\rm 1}, Huanrui Yang\textsuperscript{\rm 2}\footnotemark[1], Youxin Chen\textsuperscript{\rm 3}, Rongyu Zhang\textsuperscript{\rm 1}, Miao Wang\textsuperscript{\rm 1}, Yuan Du\textsuperscript{\rm 1,4}, Li Du\textsuperscript{\rm 1,4}\footnote{Corresponding author.}
}
\affiliations{
    \textsuperscript{\rm 1}School of Electronic Science and Engineering, Nanjing University\\ \textsuperscript{\rm 2}University of Arizona\\ \textsuperscript{\rm 3}Samsung Electronic Research Centre of China\\
    \textsuperscript{\rm 4}Interdisciplinary Research Center for Future Intelligent Chips, Nanjing University, Suzhou\\
        \{liuyijiang, rongyuzhang, wangmiao\}@smail.nju.edu.cn\\ huanruiyang@arizona.edu, yx113.chen@samsung.com,
        \{yuandu, ldu\}@nju.edu.cn
}

\usepackage{bibentry}

\begin{document}

\maketitle

\begin{abstract}
Large language models (LLMs) excel in language tasks, especially with supervised fine-tuning after pre-training. However, their substantial memory and computational requirements hinder practical applications. Structural pruning, which reduces less significant weight dimensions, is one solution. Yet, traditional post-hoc pruning often leads to significant performance loss, with limited recovery from further fine-tuning due to reduced capacity. Since the model fine-tuning refines the general and chaotic knowledge in pre-trained models, we aim to incorporate structural pruning with the fine-tuning, and propose the \texttt{Pruning-Aware Tuning} (PAT) paradigm to eliminate model redundancy while preserving the model performance to the maximum extend.
Specifically, we insert the innovative Hybrid Sparsification Modules (HSMs) between the Attention and FFN components to accordingly sparsify the upstream and downstream linear modules. The HSM comprises a lightweight operator and a globally shared trainable mask. The lightweight operator maintains a training overhead comparable to that of LoRA, while the trainable mask unifies the channels to be sparsified, ensuring structural pruning.
Additionally, we propose the Identity Loss which decouples the transformation and scaling properties of the HSMs to enhance training robustness.
Extensive experiments demonstrate that PAT excels in both performance and efficiency. For example, our Llama2-7b model with a 25\% pruning ratio achieves 1.33$\times$ speedup while outperforming the LoRA-finetuned model by up to 1.26\% in accuracy with a similar training cost. 
\end{abstract}
\begin{links}
    \link{Code}{https://github.com/kriskrisliu/PAT}
\end{links}

%

\section{Introduction}

\begin{figure}[t]
    \centering
    \resizebox{0.8\linewidth}{!}{\includegraphics{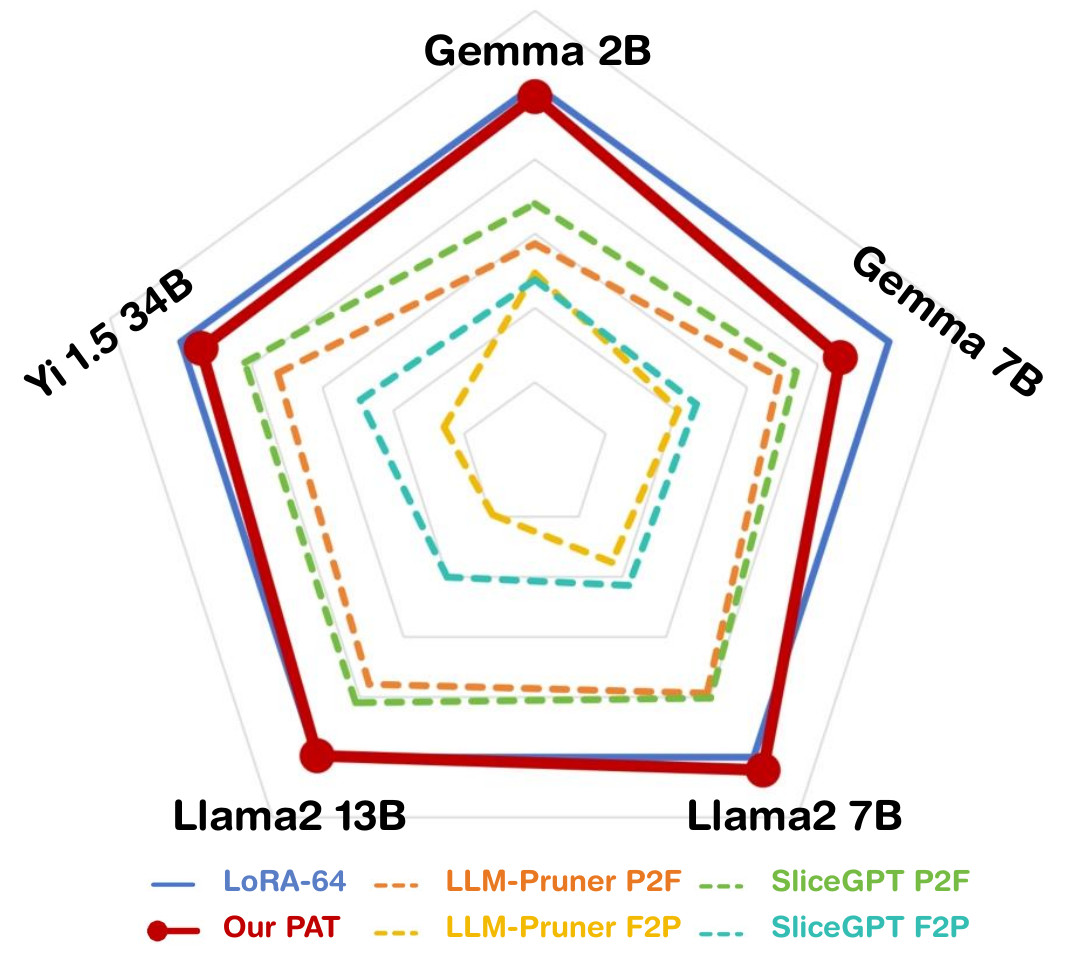}}
    \caption{Comparison of zero-shot accuracy averaged on downstream tasks. Various pruning methods at a 25\% pruning ratio, as well as the unpruned LoRA, are employed. Our PAT (red) notably outperforms LLM-Pruner and SliceGPT, and is comparable to LoRA (blue), surpassing LoRA by 1.26\% on the Llama2-7B model.}
    \label{fig:size-acc}
\end{figure}
Large language models (LLMs)~\cite{Touvron2023LLaMAOA, Brown2020LanguageMA, Chowdhery2022PaLMSL} have transformed the field of NLP~\cite{Vaswani2017AttentionIA, Bahdanau2014NeuralMT, Zhang2015CharacterlevelCN, Yang2016HierarchicalAN} with their exceptional performance on various complex language benchmarks. Despite their success, these models often necessitate substantial computational resources and present challenges for practical deployment due to their billions of parameters. Their extensive scales result in high latency and complications in deployments~\cite{Pan2023SmoothQuantAA,zhang2024efficient}. To mitigate these issues, various techniques have been proposed, including model pruning~\cite{Ma2023LLMPrunerOT,ashkboos2024slicegpt,Sun2023ASA, Santacroce2023WhatMI, Fang2023StructuralPF}, knowledge distillation~\cite{Agarwal2023OnPolicyDO, Tunstall2023ZephyrDD, Sun2019PatientKD, Sun2020ContrastiveDO, Ma2020AdversarialSD}, and quantization~\cite{Liu2022NoisyQuantNB, Yao2022ZeroQuantEA, Bai2020BinaryBERTPT, Zafrir2019Q8BERTQ8} within the context of pre-trained language models (PLMs).

Network pruning~\cite{Syed2023PruneAT, Xu2021RethinkingNP, Liu2021EBERTEB, Guo2019ReweightedPP}, which reduces model size by eliminating specific weights, has gained significant attention. Especially for structural pruning~\cite{ashkboos2024slicegpt, Li2016PruningFF, Wang2019EigenDamageSP} which promises practical acceleration on current hardware architectures. However, as shown in \cref{fig:size-acc}, traditional pruning methods~\cite{Ma2023LLMPrunerOT,ashkboos2024slicegpt} usually results in significant performance loss, whether applied before or after recovery model finetuning with Pre/Post-Trainig Pruning (P2F/F2P). 

On the other hand, since the pretraining-fine-tuning pipeline has become standard practice in both academic and industrial scenarios, Parameter-Efficient Fine-Tuning (PEFT) methods~\cite{Xu2023ParameterEfficientFM, Lin2020ExploringVG, Mahabadi2021ParameterefficientMF,liu2024intuition}, e.g., Low-Rank Adapter (LoRA)~\cite{Hu2021LoRALA}, have emerged as prevailing solutions for streamlined training. Meanwhile, since model fine-tuning can be seen as refining the universal and chaotic knowledge in the pre-trained model, thereby transforming the general LLM into a task-specific expert, combining structural pruning and PEFT for model efficiency and quick adaptation becomes a natural thought.

Drawing inspiration from quantization methods that often work synergistically, including the training-free Post-Training Quantization (PTQ)~\cite{Dettmers2022LLMint88M, Frantar2022GPTQAP, Lin2023AWQAW, Lee2023OWQLL} and the performance-enhancing Quantization-Aware Training (QAT)~\cite{Liu2023LLMQATDQ, Kim2023MemoryEfficientFO, Dettmers2023QLoRAEF}, we aim to incorporate structure pruning into the fine-tuning process while further boosting the model performance. This prompts us to introduce a new \texttt{Pruning-Aware Tuning} (PAT) paradigm to facilitate efficient inference and practical deployment in real-world applications, such as autonomous vehicles which require fast and accurate model inference to make real-time decisions and avoid obstacles while a fine-tuned RAG model must quickly and precisely retrieve and generate relevant responses from a compact knowledge base for different customer support. Unlike traditional P2F/F2P methods that remove model weights based on fixed prior knowledge, our proposed PAT method enables simultaneous pruning and fine-tuning. This allows the model to adaptively learn which parameters are most redundant and should be pruned during the PAT process. As a result, we achieve an automatic, end-to-end structured pruning process that not only maximizes but can also enhance the capabilities of the fine-tuned model.

Specifically, we propose the integration of plug-in Hybrid Sparsification Modules (HSMs). These HSMs are strategically positioned between the Attention and FFN components. Initially, they are set as identity matrices to maintain stable gradients at the onset of the fine-tuning process. As fine-tuning progresses, the HSMs selectively attenuate the channel values of the hidden dimensions, resulting in the exclusion of the corresponding linear projection weights.
However, directly integrating dense-structured HSMs introduces an excess of trainable parameters. To mitigate this issue, we leverage the Hybrid-Identity-Operator (HIO), which reduces the number of trainable parameters. Compared with other PEFT methods, our approach not only achieves parameter efficiency but also decreases the overall model complexity. Furthermore, we introduce the Identity Loss (IL) applied to the HSMs to enhance training robustness and efficacy. This technique regularizes the HSMs while delegating the scaling functionality to independent trainable parameters.

In addition, the pruning operation across all HSMs is governed by a single trainable Unified Sparsification Mask (USM), ensuring consistent retention of channel indices across modules. This approach standardizes and streamlines the transformer decoder structure. As the trainable mask gradually converges to the target sparsity, the knowledge encoded in weights from pruned channels are seamlessly updated and redistributed to the remaining active channels. 

Extensive experiments on widely recognized Large Language Models (LLMs) demonstrate the effectiveness of our proposed \texttt{Pruning-Aware Tuning} (PAT) compared to state-of-the-art baselines, including Parameter-Efficient Fine-Tuning (PEFT) and Pre/Post-Training Pruning (PTP) methods. Notably, on the Llama2-7B model, PAT surpasses the performance of LoRA-64 by 1.26\% while achieving 25\% weight pruning. The contribution of this paper can be summarized as follows:
\begin{itemize}
\item We propose an innovative paradigm called \texttt{Pruning-Aware Tuning} (PAT). Unlike traditional pre- or post-training pruning methods, PAT achieves simultaneous structural pruning and fine-tuning, leading to improved model performance.
\item To decrease overall model complexity, we integrate plug-in Hybrid Sparsification Modules (HSMs) with the Hybrid-Identity-Operator. Additionally, we design an Identity Loss (IL) applied to the HSMs to further enhance fine-tuning efficiency and robustness.
\item We utilize a single Unified Sparsification Mask (USM) that governs all HSMs, ensuring consistent retention of channel indices across modules.
\end{itemize}

\section{Related Work}
\subsection{Pruning}
Network pruning~\cite{LeCun1989OptimalBD} has long been recognized as an effective method for model compression and acceleration. Earlier research primarily focused on small-scale networks~\cite{Fang2023DepGraphTA,yang2023global,chen2021only,wu2024auto}. However, with the advent of large-scale models, pruning techniques have increasingly been applied to large language models (LLMs).
According to the pruning granularity, pruning methods can be broadly categorized into unstructured  and structured pruning. 
In the realm of unstructured pruning~\cite{Frantar2023SparseGPTML,Sun2023ASA}, techniques such as SparseGPT~\cite{Frantar2023SparseGPTML} and Wanda~\cite{Sun2023ASA} have been proposed. SparseGPT addresses the layer-wise reconstruction problem by utilizing Hessian inverses, while Wanda employs the product of weight magnitudes and input feature norms as its pruning criterion. Despite their effectiveness, these unstructured sparsification methods do not guarantee on-device speedup without hardware-specific support. 
In contrast, the structured pruning~\cite{Zafrir2021PruneOF, Kurtic2022TheOB, Xia2022StructuredPL,yang2019deephoyer,yang2023global} removes organized patterns within the network, enabling significant acceleration in a hardware-agnostic manner. For instance, Shortened-LLaMA~\cite{Kim2024ShortenedLA} removes Transformer blocks, resulting in depth pruning. Sheared-LLaMA~\cite{Xia2023ShearedLA} incorporates the learnable mask to prune both the network’s width and depth. LLM-Pruner~\cite{Ma2023LLMPrunerOT} and SliceGPT~\cite{ashkboos2024slicegpt} prune the network width while retaining the number of layers: LLM-Pruner sparsifies the intermediate dimension while SliceGPT focuses on the hidden dimension.
However, existing structured pruning models still suffer from accuracy loss, necessitating further exploration and improvement.

\subsection{Parameter-Efficient Fine-Tuning}
Compared to full fine-tuning of LLMs, Parameter-Efficient Fine-Tuning (PEFT) can achieve comparable performance while significantly reducing the computation and memory cost. PEFT methods can be broadly classified into five categories: additive fine-tuning, partial fine-tuning, reparameterized fine-tuning, hybrid fine-tuning, and unified fine-tuning. Additive fine-tuning methods introduce new additional parameters into the model, including adapter-based~\cite{Hu2021LoRALA,Zhang2023LLaMAAdapterEF, He2021TowardsAU, Rckl2020AdapterDropOT} and soft prompt-based~\cite{Li2021PrefixTuningOC, Wang2023MultitaskPT, Vu2021SPoTBF} approaches. For example, LoRA~\cite{Hu2021LoRALA}, one of the most popular used PEFT method, freezes the pre-trained model weights and injects trainable rank decomposition matrices into each layer of the Transformer architecture, greatly reducing the number of trainable parameters for downstream tasks. DoRA~\cite{liu2024dora}, a successful variant of LoRA, achieves enhanced performance by decomposing the pre-trained weights into magnitude and direction for subsequent fine-tuning. Partial fine-tuning selects only the parameters that are important for the downstream task to be trained~\cite{BenZaken2021BitFitSP, Lawton2023NeuralAS, Xu2021RaiseAC}. 
Reparameterized fine-tuning methods~\cite{Edalati2022KronAPE, Zhang2023PruningML, Xu2023QALoRAQL} often use low-rank transformations to reduce the number of trainable parameters.
Hybrid fine-tuning~\cite{Zhou2023AutoPEFTAC, Hu2022SparseSS} combines multiple PEFT methods together. Unified fine-tuning~\cite{He2022SparseAdapterAE, Wang2022AdaMixMF} integrates various fine-tuning methods into a unified structure, but only utilizes one of them during fine-tuning.
In this study, we mainly employ LoRA and DoRA as the fine-tuning techniques to explore our proposed PAT paradigm.

\section{Methodology}
In this section, we detail the components of our proposed \texttt{Pruning-Aware Tuning} (PAT). Firstly, we introduce the foundational concept of the zero-preservation property inherent in the RMSNorm operation. Subsequently, we elaborate on the Hybrid Sparsification Module (HSM) and the Unified Sparsification Mask (USM). Furthermore, we outline the comprehensive process of PAT and introduce the innovative Identity Loss (IL). Finally, we expound on the overall optimization objective.

\begin{figure*}[t]
    \centering
    \resizebox{\linewidth}{!}{\includegraphics{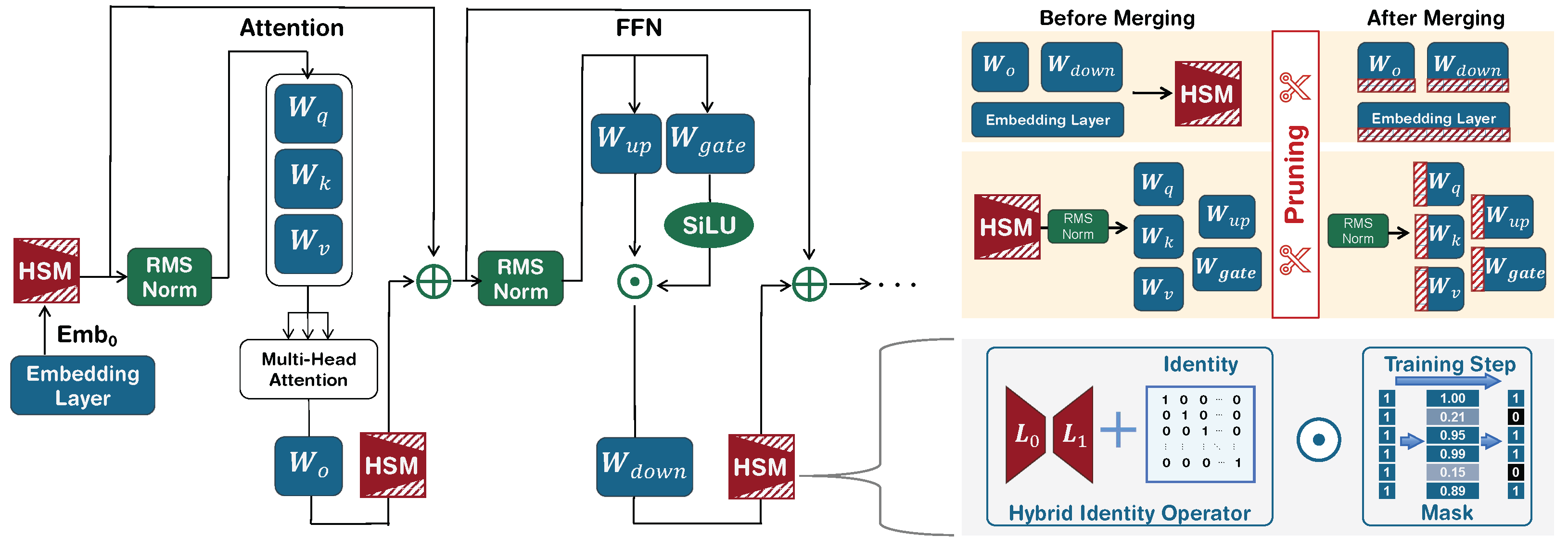}}
    \caption{Framework of our Pruning-Aware Tuning (PAT), featuring Hybrid Sparsification Modules (HSMs) positioned between the Attention and Feed-Forward Network (FFN) components. Each HSM includes a Hybrid-Identity-Operator (HIO) and a globally shared trainable mask. At training stage, the mask values will be updated until convergence. At inference stage, the pruned HSMs and the upstream linear layers will be merged, and the downstream layers which receive inputs with zero-valued channels will be pruned accordingly.}
    \label{fig:method-drm}
\end{figure*}

\subsection{Preliminary: Zero-Preservation of RMSNorm}
RMSNorm~\cite{Zhang2019RootMS}, an abbreviation for root mean square layer normalization, is widely used in LLMs, such as Llama~\cite{touvron2023llama}, Gemma~\cite{team2024gemma}, and Yi~\cite{young2024yi}. The general form of the RMSNorm is defined as the following:
\begin{equation}
    \label{eq:rms-norm}
    \Bar{x}_{i} = \operatorname{RMSNorm}(x_i) = \dfrac{x_{i}}{\operatorname{RMS}(\mathbf{x})} g_{i},
\end{equation}
where $\Bar{x}_i$ is the $i$-th value of vector $\Bar{\mathbf{x}}\in \mathds{R}^d$, and
$\mathbf{g}\in \mathds{R}^d$ is the gain parameter. $\operatorname{RMS}(\cdot)$ is the Root Mean Square operation, defined as:
\begin{equation}
    \operatorname{RMS}(\mathbf{x}) = \sqrt{\dfrac{1}{d}\sum_{i=1}^{d}x_{i}^{2}}
\end{equation}
Given the layer input $\mathbf{X}\in \mathds{R}^{d\times n}$ with specific (e.g., 1st and 2nd) channels all equal to $\mathbf{0}$ :
\begin{equation}
\mathbf{X} =\begin{pmatrix}
0 & 0 & \cdots & 0 \\
0 & 0 & \cdots & 0 \\
x^{(1)}_3 & x^{(2)}_3 & \cdots & x^{(n)}_3 \\
\vdots & \vdots & \ddots & \vdots \\
x^{(1)}_d & x^{(2)}_d & \cdots & x^{(n)}_d \\
\end{pmatrix}
\end{equation}
where $x^{(i)}_{j}$ is the $j$-th value of the $i$-th vector in $\mathbf{X}$.
Referring to \cref{eq:rms-norm}, the RMSNorm operation will preserve these zero values, thereby making it feasible to prune the corresponding channels.

\subsection{Hybrid Sparsification Module (HSM)}
Our objective is to prune the hidden dimensions of LLMs during fine-tuning, which would involve selecting the channels to be pruned in a linear layer, and convert the knowledge of pruned weights into those remained. To achieve this, we design a specific module to be applied after a linear layer, namely Hybrid Sparsification Module (HSM). HSM consists of a trainable channel selection mask $\mathbf{M}$ and a knowledge transformation weight $\mathbf{D}$. Specifically, the computation involving the HSM and the upstream linear layer with weight $\mathbf{W}\in \mathds{R}^{d_{o} \times d_i}$ is formulated as follows:
\begin{equation}
\label{equ:HSM}
    \begin{aligned}
        \mathbf{Z} &= (\mathbf{M} \odot \mathbf{D})\cdot \mathbf{W} \mathbf{X}\\
        &= (\mathbf{M} \odot \mathbf{D W}) \cdot \mathbf{X}\\
        &= \mathbf{W}_D \cdot \mathbf{X},
    \end{aligned}
\end{equation}
where $d_i$ and $d_o$ are the input and output dimension, respectively, $\mathbf{X}\in \mathds{R}^{d_{i}\times n}$ is the input value, $\mathbf{Z}\in \mathds{R}^{d_o \times n}$ is the output value, $\mathbf{M}\in \mathds{R}^{d_o}$ denotes the trainable mask whose values converge to either 0 or 1, $\mathbf{D} \in \mathds{R}^{d_o \times d_o}$ is the HSM weight, $\mathbf{W}\in \mathds{R}^{d_o\times d_i}$ is the upstream linear weight, and $\mathbf{W}_D\in \mathds{R}^{d_{o} \times d_i}$ is the merged weight that  replaces $\mathbf{W}$ after training. Notably, the zero values in $\mathbf{M}$ effectively cause the corresponding output channels of $\mathbf{W}_D$ to be pruned.

To prune all linear layers in LLMs such as Llama2, which encompass the Q, K, V, and O projections in Attentions, as well as Up, Gate, and Down projections in FFNs, a straightforward approach is to apply the HSM after all linear layers. However, considering the sheer number of the linear layers in an LLM, this approach would incur significant overhead. We propose a novel and efficient alternative: placing pruning modules only between the Attention and FFN components, as illustrated in \cref{fig:method-drm}. 
The ``pruned\footnote{At this point, we indicate the zero-valued channels as pruned ones to explain the feasibility of pruning in downstream computations.}'' HSM's output, $\mathbf{Z}$, will first undergo the addition with the residual connection, which has already been pruned by the previous HSM, and then be fed into the RMSNorm operator before the next Attention/FFN component. As demonstrated previously in the preliminary, the RMSNorm has no impact on zero-valued channels, and since the downstream linear projection receives input with certain channels set to zero, the input dimensions of the following block can be pruned accordingly.
In cases where LLMs involve the LayerNorm which projects zero-valued channels to non-zero, we can convert it to the RMSNorm before incorporating HSMs. This transformation is mathematically equivalent, as described by SliceGPT~\cite{ashkboos2024slicegpt}.

Although inserting HSMs between Attention and FFN components reduces trainable parameters compared to directly applying them to each linear module, the overall training overhead remains significantly larger than that of PEFT methods. To mitigate this issue, we propose the Hybrid-Identity-Operator (HIO) as a replacement for the dense structure of HSMs, which is formulated as: 
\begin{equation}
\label{eq:drm-w/o-IL}
    \mathbf{D} = \mathbf{L}_{1}\cdot \mathbf{L}_{0} + \mathbf{I},
\end{equation}
where $\mathbf{L}_0 \in \mathds{R}^{r \times d_o}$, $\mathbf{L}_1 \in \mathds{R}^{d_o \times r}$, $r$ is the rank value of $\mathbf{L}_{1} \mathbf{L}_{0}$, and $\mathbf{I} \in \mathds{R}^{d_o \times d_o}$ is the identity matrix with diagonal values set to 1 and other values set to 0. During fine-tuning, $\mathbf{I}$ is frozen, allowing gradients to flow through $\mathbf{L}_0$ and $\mathbf{L}_1$.
HIO significantly reduces the number of trainable parameters.
For example, a dense HSM consists of $d_o \times d_o$ parameters, while the HIO consists of $2\times d_o \times r$. By determining $r<d_o/2$, we can decrease the number of trainable parameters. In practice, we set $r$ to approximately 5\% of $d$, which in turn only accounts for 10\% parameter of dense HSMs. 

\subsection{Unified Sparsification Mask (USM)}
We utilize a single trainable mask $M$ as in~\cref{equ:HSM} to adaptively set channel values of hidden states to zero. The mask acts uniformly across all HSMs, ensuring consistency in the pruned channel indices throughout the computation flow. This unified pruning mask is particularly necessary at the residual connections between Attention and FFN components, as it guarantees that the pruned channels are correctly aligned throughout the entire data flow. 

\begin{figure}[t]
    \centering
    \resizebox{\linewidth}{!}{\includegraphics{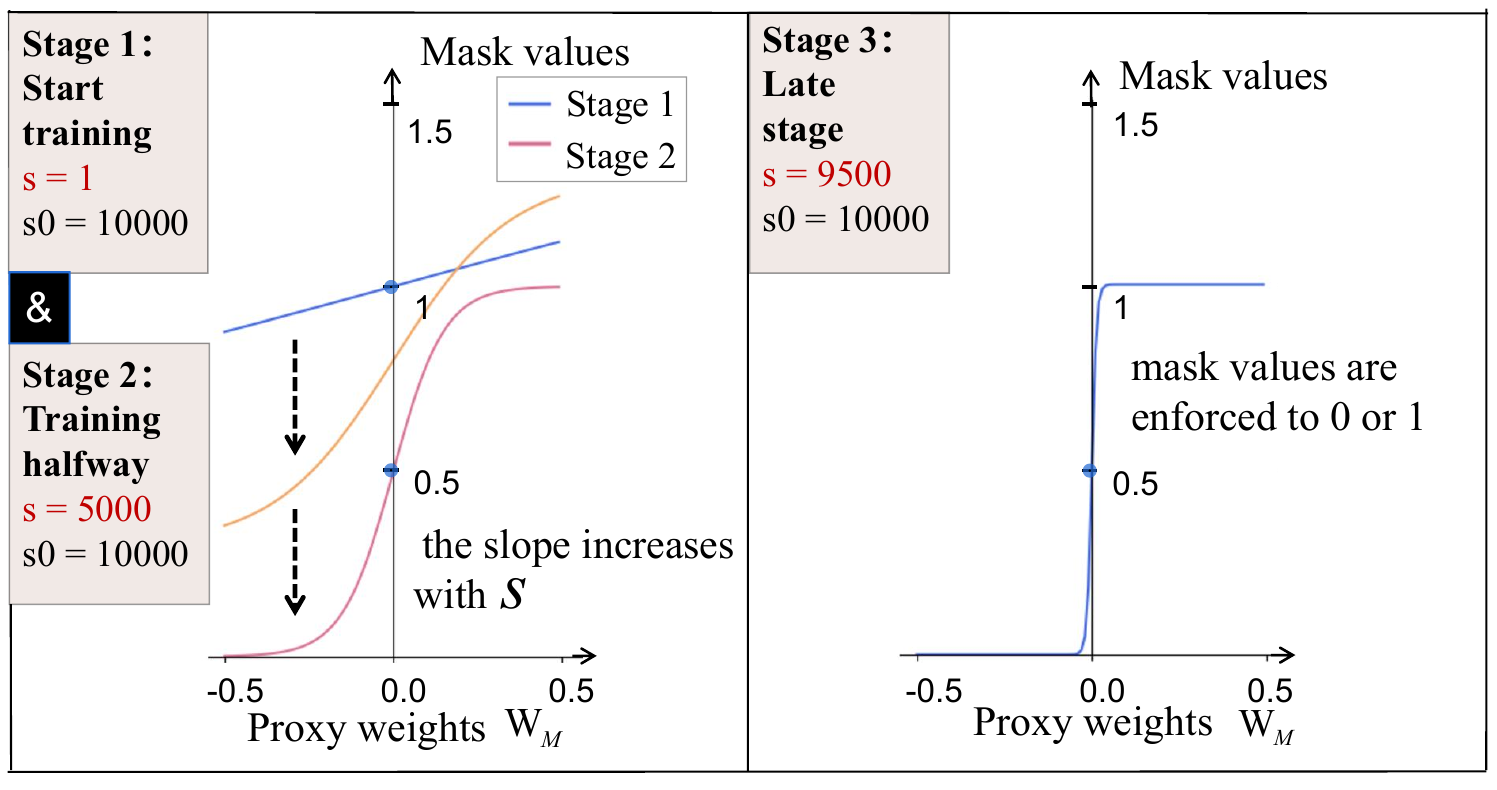}}
    \caption{The differentiable gating function $\mathcal{G}(\cdot)$.}
    \label{fig:method-temperature}
\end{figure}

To insure structural sparsity at the convergence of the model, we employ a continuous sparsification strategy with a tailored regularizer to ensure that the mask converges to discrete values of 0 or 1 and achieves the desired sparsity at the end of the training process. This involves applying a differentiable gating function, $\mathcal{G}(\cdot)$, to the trainable proxy weights $\mathbf{W}_M$ of the mask. The gating function utilizes a modified Sigmoid function with a variable temperature $\tau$, which is defined as:
\begin{equation}
    \tau(s) =
    \begin{cases}
        \dfrac{1}{1-\dfrac{\ln(s)}{\ln(s_0)}} & \text{if } s<s_0, \\
        \\
        \epsilon^{-1} & \text{otherwise.}
    \end{cases}
\end{equation}
\begin{equation}
    \beta(s) = 
    \begin{cases}
        \dfrac{-s}{s_0} + 0.5 & \text{if } s < s_{0}/2, \\
        \\
        0 & \text{otherwise.}
    \end{cases}
\end{equation}
\begin{equation}
    \mathbf{M} = \mathcal{G}(s,\mathbf{W}_{M}) = \dfrac{1}{1+e^{-\tau(s) \cdot \mathbf{W}_{M}}} + \beta{(s)},
\end{equation}
where $s$ denotes the current training step which dynamically determines the temperature, $s_0$ is the milestone step which indicates that the temperature stay unchanged in the remaining training steps. In practice, we set $s_0$ to $1/3$ of the total training steps. $\beta(\cdot)$ denotes the offset which varies according to the step. \cref{fig:method-temperature} demonstrates some typical training stages. 
Initially, when $s=0$, the gating function maps all proxy weights of the mask to 1. This is achieved by initializing $\mathbf{W}_M$ to zero, which keeps the model weights unchanged, ensuring stable gradients at the beginning. As the temperature increases, the slope near 0 rises, and the offset term decreases. By halfway to the milestone step, the offset term reaches 0 and stops updating, while the slope continues to increase. At the milestone step, the slope near 0 becomes very steep, while the slope elsewhere approaches 0. At this point, the mask values will be enforced to either 0 or 1, where 0 refers to the channel being pruned.
Moreover, to achieve the target sparsity, specifically the proportion of values equal to 0, we propose regularizing the number of active channels. This is achieved through the following regularization term:
\begin{equation}
    \mathcal{L}_{active} = \|N_{target}-\sum_{i}{\mathds{1}_{(m_i>0)}}\|_2,
\end{equation}
where $N_{target}$ denotes the target channel number of active channels, $m_i$ represents the $i$-th value of the proxy weight $M$, and $\mathds{1}_{(condition)}$ is the indicator function that equals 1 if the condition is true, and 0 otherwise.

\subsection{Pruning-Aware Tuning} %
We perform model fine-tuning by updating the proposed HSM modules and applying LoRA on all linear layers~\cite{Hu2021LoRALA}.
Besides the standard instruction fine-tuning loss $\mathcal{L}_{Instruct}$,
we propose the innovative Identity Loss (IL) to decompose the scaling and rotation in the HSM transformations.
Specifically, we alter the formulation of~\cref{eq:drm-w/o-IL} into:
\begin{equation}
    \mathbf{D} = \mathbf{L}_{1} \cdot \operatorname{diag}(\mathbf{v}) \cdot \mathbf{L}_{0} + \mathbf{I},
\end{equation}
where $\mathbf{v} \in \mathds{R}^{r}$ is the trainable scaling values, and $L_0$ and $L_1$ are constrained to be orthogonal with the identity regularization
\begin{equation}
    \mathcal{L}_{Identity} = \|\mathbf{L}_{0}\cdot \mathbf{L}^{T}_{0} - \mathbf{I}\|_2 + \|\mathbf{L}^{T}_{1}\cdot \mathbf{L}_{1} - \mathbf{I}\|_2
\end{equation}
The overall optimization objective is defined by a composite loss function $\mathcal{L}$, which is expressed as follows:
\begin{equation}
    \mathcal{L} = \mathcal{L}_{Instruct} + \mathcal{L}_{active} + \mathcal{L}_{Identity},
\end{equation}
where $\mathcal{L}_{Instruct}$ represents the loss associated with instruction fine-tuning.

\section{Experiments}
In this section, we present the experimental results and analysis. We begin by describing the experimental setup. Next, we showcase our main results across various Language Models (LLMs). We then delve into the efficiency and accuracy trade-off, examining memory and latency considerations. Finally, we conduct ablation studies on the trainable mask and identity loss.
\subsection{Experimental Setup}
\paragraph{Models.}
We utilize model frameworks and checkpoints from HuggingFace~\cite{jain2022huggingface,wolf2019huggingfaceTrans}, which includes Llama-2 7B and 13B~\cite{touvron2023llama}, Gemma 2B and 7B~\cite{team2024gemma}, Yi-1.5-34B~\cite{young2024yi}.
\paragraph{Baselines.}
The pruning baselines include LLM-Pruner~\cite{Ma2023LLMPrunerOT}, and SliceGPT~\cite{ashkboos2024slicegpt}. We also involve the common LoRA~\cite{Hu2021LoRALA} approach with the rank set to 64. Unless otherwise stated, we adjust the number of trainable parameters in all fine-tuning approaches to match the number of the LoRA.
Additionally, we conduct complementary tests by applying ``P$\rightarrow$\text{FT}" (Pruning before Fine-Tuning) and ``FT$\rightarrow$\text{P}" (Fine-Tuning before Pruning) strategies on LLM-Pruner and SliceGPT. The pruning ratios are set to 20\%, 25\%, and 30\%, respectively.
\paragraph{Datasets.}
We employ the LaMini-instruction dataset~\cite{wu2023lamini} for fine-tuning. To reduce training costs, we randomly drop 50\% of the samples, resulting in a final dataset of 1 million samples. Unless otherwise stated, all experimental results are based on this setting.
We conduct zero-shot evaluation on 14 datasets, including ARC-Challenge~\cite{allenai:arc}, ARC-Easy~\cite{allenai:arc}, BOOLQ~\cite{wang2019superglue}, COPA~\cite{wang2019superglue}, HellaSwag~\cite{zellers2019hellaswag}, MMLU~\cite{hendryckstest2021}, MultiRC~\cite{wang2019superglue}, OpenBookQA~\cite{OpenBookQA2018}, PIQA~\cite{Bisk2020}, RTE~\cite{wang2019superglue}, SIQA~\cite{sap2019socialiqa}, WIC~\cite{wang2019superglue}, WinoGrande~\cite{ai2:winogrande}, WSC~\cite{wang2019superglue}. The accuracy is calulated by First-Capital-Word\footnote{https://github.com/open-compass/opencompass}~\cite{2023opencompass} method.

\paragraph{Implementation Details.}
Experiments are conducted using A100 GPUs. The models are fine-tuned over 3 epochs using the Alpaca instruction template. The learning rate is set to $5 \times 10^{-5}$ with a cosine schedule. The batch size is set to 128, and the sequence length is 256 tokens. The milestone step of our PAT, $s_0$, is set to $1/3$ of the total training steps. The settings of our HIOs are derived to match the number of trainable parameters with LoRA-64. For example, we set the rank values of HIO and LoRA modules to 200 and 20 in the Llama2-7B experiments, respectively.

\subsection{Experimental Results and Analysis}
\begin{figure}
    \centering
    \includegraphics[width=\linewidth]{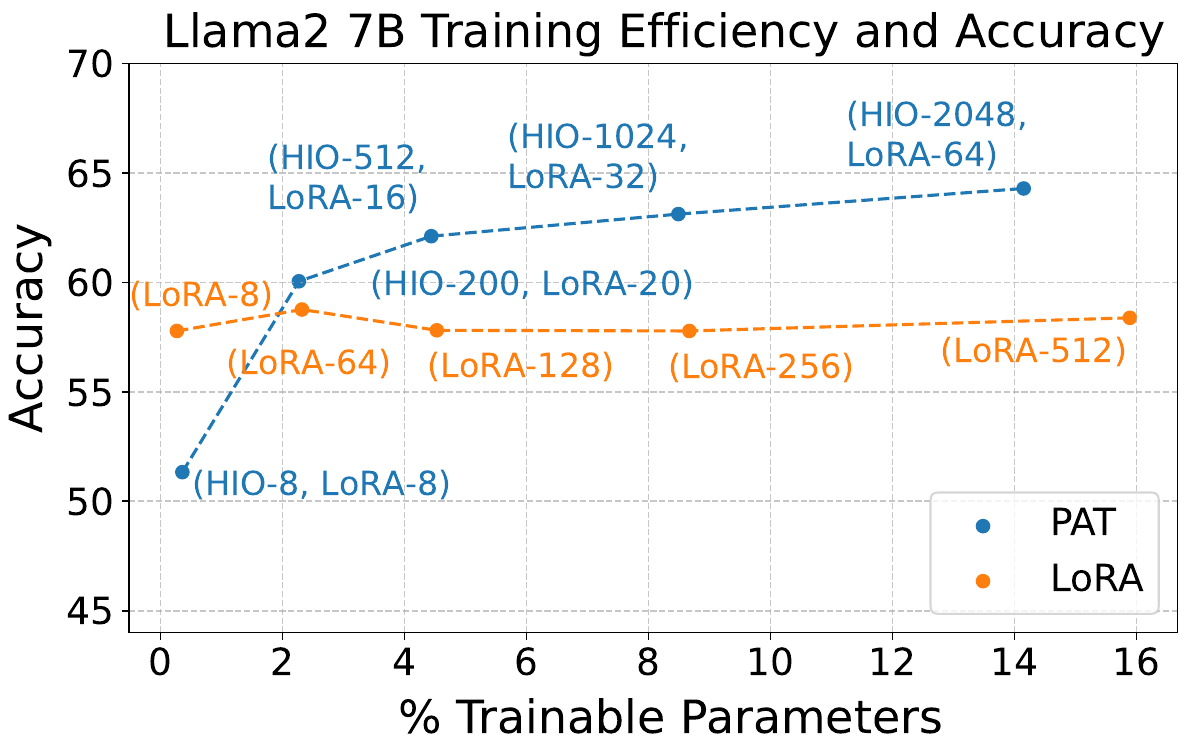}
    \caption{The training efficiency and the accuracy comparison for Llama2 7B. Our PAT results are represented as ``HIO-M, LoRA-N'', where M and N denote the rank value in the HIO and the LoRA, respectively. The LoRA results are ``LoRA-N''. }
    \label{fig:tradeoff-param-acc}
\end{figure}

\begin{figure}
    \centering
    \begin{subfigure}[b]{0.49\linewidth} %
        \centering
        \includegraphics[width=\linewidth]{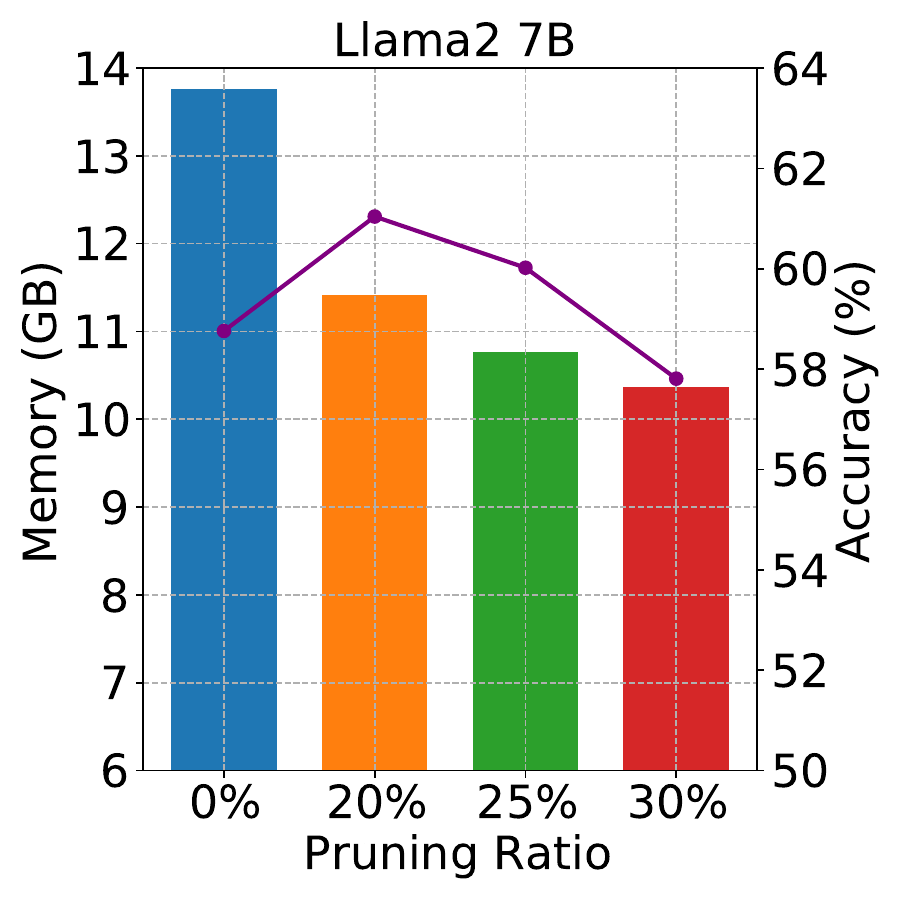}
        \caption{Llama 2 7B}
        \label{fig:subfig1}
    \end{subfigure}
    \hfill %
    \begin{subfigure}[b]{0.49\linewidth} 
        \centering
        \includegraphics[width=\linewidth]{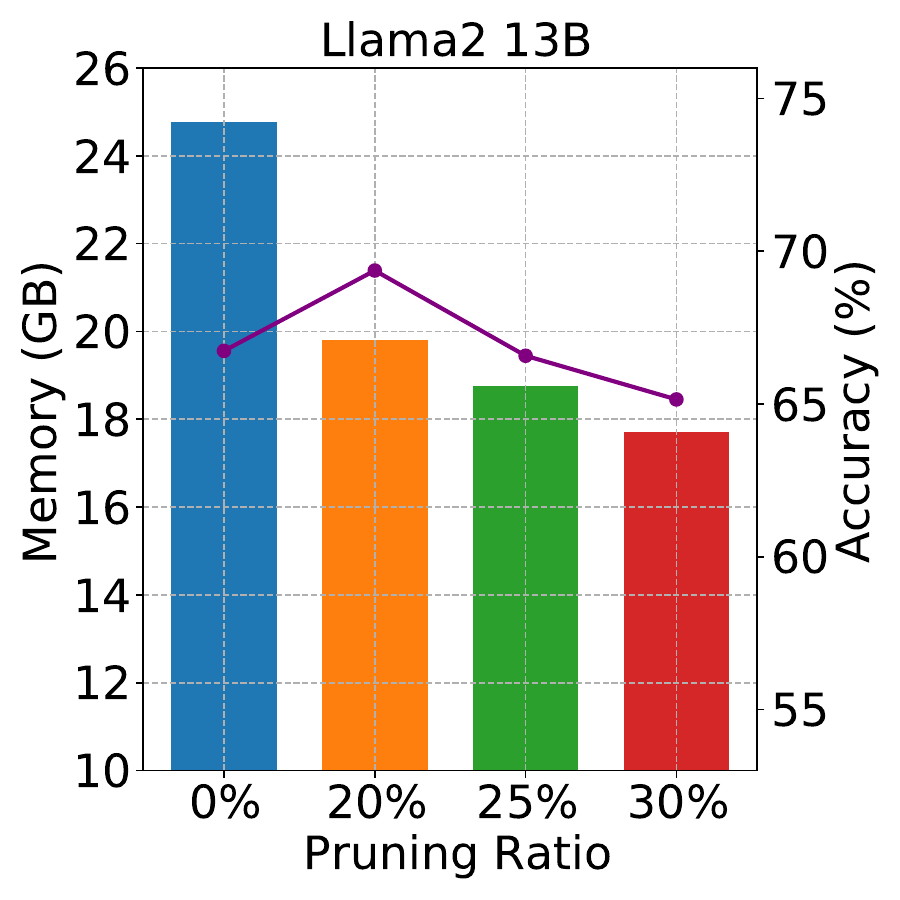}
        \caption{Llama 2 13B}
        \label{fig:subfig2}
    \end{subfigure}
    \caption{The VRAM usage and the evaluation accuracy of Llama2 models under various pruning ratios.}
    \label{fig:memory}
\end{figure}

\begin{figure}
    \centering
    \begin{subfigure}[b]{0.49\linewidth} %
        \centering
        \includegraphics[width=\linewidth]{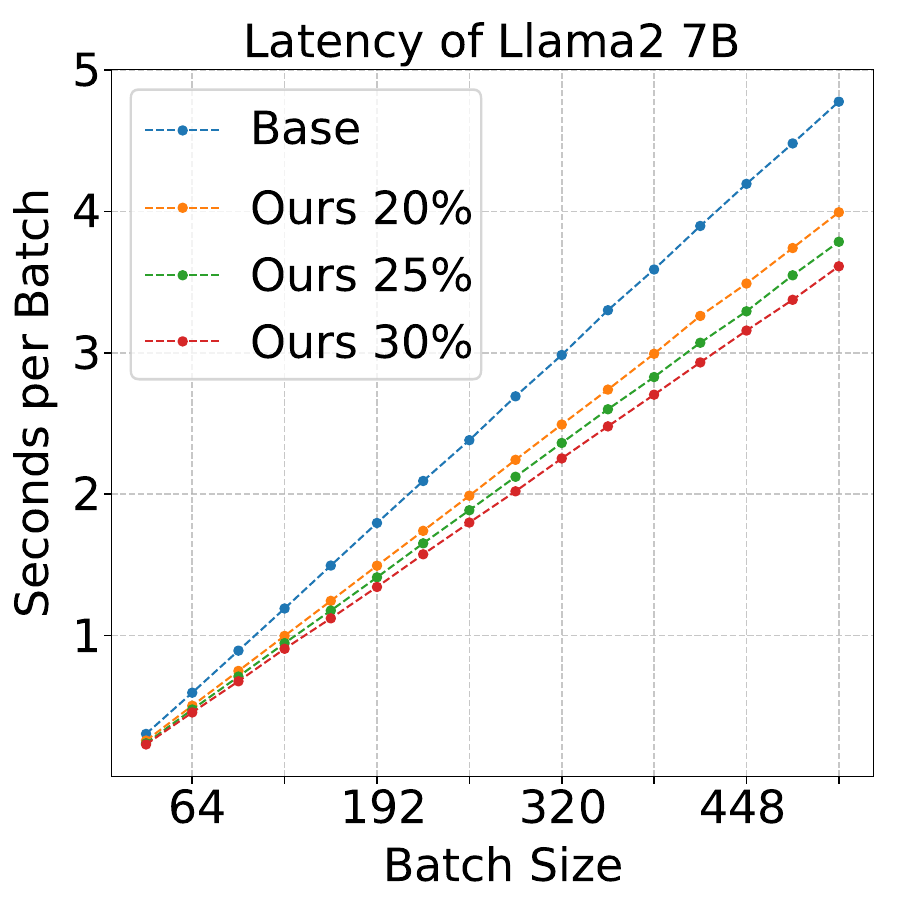}
        \caption{Llama 2 7B}
    \end{subfigure}
    \hfill %
    \begin{subfigure}[b]{0.49\linewidth} %
        \centering
        \includegraphics[width=\linewidth]{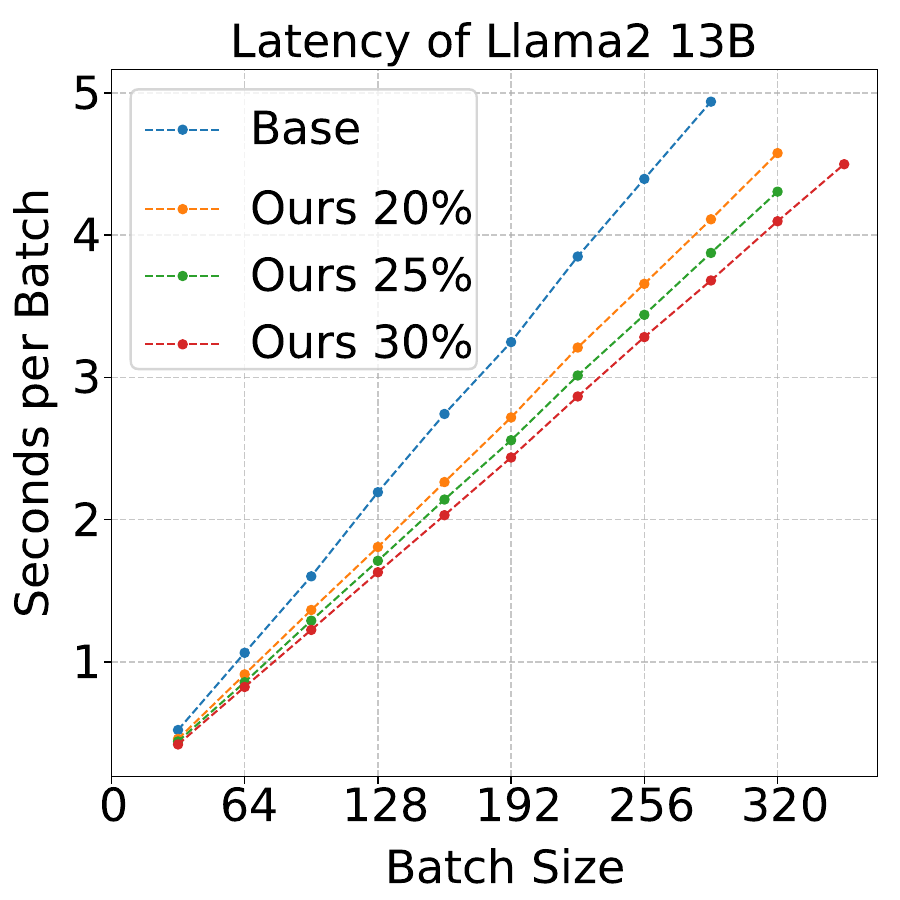}
        \caption{Llama 2 13B}
    \end{subfigure}
    \caption{The speedup of Llama2 models according to different pruning ratios and batch sizes.}
    \label{fig:latency}
\end{figure}

\paragraph{Performance Comparison.}
\begin{table*}[ht]
\centering
   	\resizebox{2\columnwidth}{!}{%
          \setlength{\tabcolsep}{5mm}{
\begin{tabular}{@{}ccc|ccccc@{}}
\toprule
\textbf{Ratio}                  & \textbf{Method}                       & \textbf{Mode}                       & \textbf{Gemma-2B}        & \textbf{Gemma-7B}        & \textbf{Llama2-7B}                & \textbf{Llama2-13B}               & \textbf{Yi-1.5-34B}               \\ \midrule
\textbf{0\%}                    & \textbf{LoRA-64}                      & FT                              &      53.82               &       71.59              &     58.76                         &    66.74                          &     81.21                      \\ \midrule
                                & LLM-Pruner                   & P$\rightarrow$FT                &      48.87               &       65.45              &      58.53                        &   65.28                           &   73.86                           \\
                                & LLM-Pruner                   & FT$\rightarrow$P                &      40.64               &       54.87              &      40.68                        &   41.43                           &   53.88                           \\
                                & SliceGPT                     & P$\rightarrow$FT                &      48.21               &       66.60              &     57.81                         &   65.86                           &   76.81                           \\
                                & SliceGPT                     & FT$\rightarrow$P                &      41.67               &       56.17              &     47.77                         &   50.67                           &   67.60                           \\
\multirow{-5}{*}{\textbf{20\%}} & \cellcolor[HTML]{FFEEAD}\textbf{Ours} & \cellcolor[HTML]{FFEEAD}\textbf{PAT}                             & \cellcolor[HTML]{FFEEAD}\textbf{53.95}           &   \cellcolor[HTML]{FFEEAD}\textbf{68.68}         & \cellcolor[HTML]{FFEEAD}\textbf{61.04}                    & \cellcolor[HTML]{FFEEAD}\textbf{69.37}                    & \cellcolor[HTML]{FFEEAD}\textbf{81.02}                     \\ \midrule
                                & LLM-Pruner                   & P$\rightarrow$FT                &      42.32               &       60.50              &      52.50                        &     58.64                         &    70.10                          \\
                                & LLM-Pruner                   & FT$\rightarrow$P                &      40.20               &       50.29              &      39.72                        &     39.82                         &    51.03                          \\
                                & SliceGPT                     & P$\rightarrow$FT                &      45.23               &       62.22              &     52.98                         &     60.69                         &   73.88                           \\
                                & SliceGPT                     & FT$\rightarrow$P                &      39.72               &       52.13              &     41.97                         &  46.75                            &   60.63                           \\
\multirow{-5}{*}{\textbf{25\%}} & \cellcolor[HTML]{FFEEAD}\textbf{Ours}                         & \cellcolor[HTML]{FFEEAD}\textbf{PAT}                             & \cellcolor[HTML]{FFEEAD}\textbf{52.98}           &  \cellcolor[HTML]{FFEEAD}\textbf{66.68}          & \cellcolor[HTML]{FFEEAD}\textbf{60.02}                    & \cellcolor[HTML]{FFEEAD}\textbf{66.58}                    & \cellcolor[HTML]{FFEEAD}\textbf{78.90}                 \\ \midrule
                                & LLM-Pruner                   & P$\rightarrow$FT                &      39.71               &       50.28              &   50.60                           &    51.28                          &   66.85                           \\
                                & LLM-Pruner                   & FT$\rightarrow$P                &      40.05               &       41.35              &    39.73                          &    39.70                          &   45.36                           \\
                                & SliceGPT                     & P$\rightarrow$FT                &      40.07               &       53.14              &    50.91                          &    56.12                          &   71.81                           \\
                                & SliceGPT                     & FT$\rightarrow$P                &      39.89               &       44.30              &    40.14                          &  46.19                            &   56.10                           \\
\multirow{-5}{*}{\textbf{30\%}} & \cellcolor[HTML]{FFEEAD}\textbf{Ours} & \cellcolor[HTML]{FFEEAD}\textbf{PAT}                             & \cellcolor[HTML]{FFEEAD}\textbf{45.33}           &   \cellcolor[HTML]{FFEEAD}\textbf{64.58}          & \cellcolor[HTML]{FFEEAD}\textbf{57.81}                    & \cellcolor[HTML]{FFEEAD}\textbf{65.15}                    & \cellcolor[HTML]{FFEEAD}\textbf{77.89}                 \\ \bottomrule
\end{tabular}}
}
\caption{Zero-shot evaluations of different pruning methods with 20\%, 25\%, and 30\% pruning ratios across various LLMs. ``\textbf{FT}'' represents \textbf{F}ine-\textbf{T}uning. ``\textbf{P$\rightarrow$FT}'' denotes \textbf{P}runing the base model and then \textbf{F}ine-\textbf{T}uning the pruned model via LoRA. ``\textbf{FT$\rightarrow$P}'' denotes \textbf{F}ine-\textbf{T}uning the base model via LoRA and then \textbf{P}runing the fine-tuned model. ``\textbf{PAT}'' denotes our proposed \textbf{P}runing-\textbf{A}ware \textbf{T}uning strategy. The accuracy is averaged across 14 datasets. More details are available in the Appendix.}
\label{tab:main-result}
\end{table*}
\cref{tab:main-result} shows the zero-shot evaluations of different pruning methods across 14 well-known tasks, where various types and sizes of LLMs are tested.
We obtain that: (1) Our method, employing the Pruning-Aware Tuning (PAT) strategy, achieves the highest accuracy across pruned models. In contrast, LLM-Pruner and SliceGPT, which use either the Pruning before Fine-Tuning (P$\rightarrow$FT) or Fine-Tuning before Pruning (FT$\rightarrow$P), suffer from non-negligible accuracy degradation. However, the``P$\rightarrow$FT'' significantly outperforms the ``FT$\rightarrow$P''.
(2) The feasibility of pruning varies across different models. We observe that Llama2 with PAT maintains comparable performance to the un-pruned LoRA approach even at a 30\% pruning rate, whereas Gemma 7b shows the trending of accuracy degradation at a 20\% pruning rate.
(3) Surprisingly, Llama2 7B and 13B with PAT under less than 30\% and 20\% pruning ratio, respectively, exhibit accuracy better than the un-pruned LoRA.
\paragraph{Efficiency and Accuracy Trade-off.}
The implementation of HIO significantly reduces the number of trainable parameters, but this reduction may directly impact the model accuracy. We conducted experiments using various scales of training parameters on the Llama 2 7B model, and illustrate the results in \cref{fig:tradeoff-param-acc}. The total number of trainable parameters is adjusted by the rank values of HIO and LoRA modules. For example, the ``LoRA-64'' in dark represents the traditional LoRA fine-tuning with a rank value set to 64, and the ``HIO-200, LoRA-20'' in purple represents our PAT with a rank of 200 in HIO and a rank of 20 in LoRA modules. We find that our PAT demonstrates a performance trend correlated to the number of trainable parameters. ``Dense\footnote{Indicating that we use the dense matrix instead of the HIO.}, LoRA-8'' with 14.15\% trainable parameters achieves 64.19\% accuracy, outperforming ``LoRA-64'' by 5.43\%. Conversely, ``HIO-8, LoRA-8'' with merely 0.36\% trainable parameters results in a 6\% accuracy reduction.
In practice, we opt for ``HIO-200, LoRA-20'' in Llama 2 7B experiments, aligning the parameter count with that of ``LoRA-64''. For others, Gemma 2B with ``HIO-300, LoRA20'', Gemma 7B with ``HIO-300, LoRA20'', Llama2 13B with ``HIO-200, LoRA20''
, and Yi-1.5 34B with ``HIO-200, LoRA20''.
\paragraph{Memory and Latency.} We conducted an evaluation of the VRAM usage and the inference latency comparing the base Llama2 7B and 13B models with pruned versions, as illustrated in \cref{fig:memory} and \cref{fig:latency}. The GPU memory is tested by loading the model without any proceeding tokens. The latency is tested by the time of the first token prediction in a batch with an initial context length of 128. Specifically, we assessed the models pruned at 20\%, 25\%, and 30\% ratios across various batch sizes. Our 30\% pruned models achieve 1.33$\times$ speedup on average. Moreover, the base Llama2 13B model encounters Out-Of-Memory (OOM) errors at a batch size of larger than 288 when executed on a single A100-80GB GPU. In contrast, our pruned models work reliably under these conditions.

\paragraph{Trainable and Frozen Mask.} The frozen mask is implemented by linearly attenuating a fixed portion of the mask values during training. In our experiment, this attenuation is applied to the first $N$ values of the hidden dimension in LLMs, where $N$ is determined by the pruning ratio. The results presented in \cref{tab:frozen-mask} demonstrate the significant advantage of the trainable mask over the frozen counterpart. For instance, in the case of the Llama2 13B model with 30\% pruning, the trainable mask yields an accuracy improvement of 4.06\% over the frozen mask.
\begin{table}[ht]
\resizebox{\linewidth}{!}{
\begin{tabular}{@{}c|ccccc@{}}
\toprule
\textbf{Model}                                                                 & \textbf{Ratio}        & \textbf{Method}      & \textbf{\begin{tabular}[c]{@{}c@{}}Trainable\\ Mask\end{tabular}} & \textbf{\begin{tabular}[c]{@{}c@{}}Identity\\ Loss\end{tabular}} & \textbf{Accuracy}        \\ \midrule
\multirow{7}{*}{\textbf{\begin{tabular}[c]{@{}c@{}}Llama2\\ 7B\end{tabular}}}  & 0\%                   & LoRA                 & N/A                                                               & N/A                                                              & 58.76                \\ \cmidrule(l){2-6} 
                                                                               & \multirow{3}{*}{25\%} & \multirow{3}{*}{PAT} & \textcolor{red}{\ding{55}}                                                            & \textcolor{rgb:red,0.0;green,0.8;blue,0.2}{\ding{51}}                                                               & 54.97                     \\
                                                                               &                       &                      & \textcolor{rgb:red,0.0;green,0.8;blue,0.2}{\ding{51}}                                                         & \textcolor{red}{\ding{55}}                                                              & 58.62            \\
                                                                               &                       &                      & \textcolor{rgb:red,0.0;green,0.8;blue,0.2}{\ding{51}}                                                & \textcolor{rgb:red,0.0;green,0.8;blue,0.2}{\ding{51}}                                                      & \textbf{60.02}       \\ \cmidrule(l){2-6} 
                                                                               & \multirow{3}{*}{30\%} & \multirow{3}{*}{PAT} & \textcolor{red}{\ding{55}}                                                            & \textcolor{rgb:red,0.0;green,0.8;blue,0.2}{\ding{51}}                                                               & 52.72                     \\
                                                                               &                       &                      & \textcolor{rgb:red,0.0;green,0.8;blue,0.2}{\ding{51}}                                                         & \textcolor{red}{\ding{55}}                                                              & 56.59            \\
                                                                               &                       &                      & \textcolor{rgb:red,0.0;green,0.8;blue,0.2}{\ding{51}}                                                & \textcolor{rgb:red,0.0;green,0.8;blue,0.2}{\ding{51}}                                                      & \textbf{57.81}       \\ \midrule
\multirow{7}{*}{\textbf{\begin{tabular}[c]{@{}c@{}}Llama2\\ 13B\end{tabular}}} & 0\%                   & LoRA                 & N/A                                                              & N/A                                                              & 66.74                \\ \cmidrule(l){2-6} 
                                                                               & \multirow{3}{*}{25\%} & \multirow{3}{*}{PAT} & \textcolor{red}{\ding{55}}                                                            & \textcolor{rgb:red,0.0;green,0.8;blue,0.2}{\ding{51}}                                                               & 62.35                \\
                                                                               &                       &                      & \textcolor{rgb:red,0.0;green,0.8;blue,0.2}{\ding{51}}                                                         & \textcolor{red}{\ding{55}}                                                             & 65.81 \\
                                                                               &                       &                      & \textcolor{rgb:red,0.0;green,0.8;blue,0.2}{\ding{51}}                                                & \textcolor{rgb:red,0.0;green,0.8;blue,0.2}{\ding{51}}                                                      & \textbf{66.58}       \\ \cmidrule(l){2-6} 
                                                                               & \multirow{3}{*}{30\%} & \multirow{3}{*}{PAT} & \textcolor{red}{\ding{55}}                                                            & \textcolor{rgb:red,0.0;green,0.8;blue,0.2}{\ding{51}}                                                               & 61.09                \\
                                                                               &                       &                      & \textcolor{rgb:red,0.0;green,0.8;blue,0.2}{\ding{51}}                                                         & \textcolor{red}{\ding{55}}                                                              & 64.85 \\
                                                                               &                       &                      & \textcolor{rgb:red,0.0;green,0.8;blue,0.2}{\ding{51}}                                                & \textcolor{rgb:red,0.0;green,0.8;blue,0.2}{\ding{51}}                                                      & \textbf{65.15}       \\ \bottomrule
\end{tabular}
}
\caption{Ablation study on trainable mask and identity loss.}
\label{tab:frozen-mask}
\end{table}

\paragraph{Ablation on Identity Loss.}
The incorporation of Identity Loss contributes to an enhanced accuracy improvement. As depicted in \cref{tab:frozen-mask}, Llama2 7B achieves 1.4\% enhancement with the pruning ratio of 25\%.
\paragraph{Downstream Task Capability.} Following the downstream task adaptation detailed in  DoRA~\cite{liu2024dora}, we leverage PAT to fine-tune on specific tasks, including ARC, SuperGlue, OpenBookQA, PIQA, SIQA, MMLU, and WinoGrande. The setting of HSMs is ``HIO-200, LoRA/DoRA-20''. Our 25\% pruned PAT-L and PAT-D achieve performance levels on par with those achieved by traditional DoRA and LoRA, shown in \cref{tab:dora}.
\begin{table}[h]
\centering
   	\resizebox{\columnwidth}{!}{%
          \setlength{\tabcolsep}{5mm}{
\begin{tabular}{@{}cc|c|c@{}}
\toprule
\textbf{Method} & \textbf{Ratio} & \multicolumn{1}{l|}{\textbf{Llama2 7B}} & \multicolumn{1}{l}{\textbf{Llama2 13B}} \\ \midrule
LoRA-64   & 0\%            & 72.85                              & 76.24                             \\
\textbf{PAT-L}   & 25\%            &  72.05                            &   76.08                           \\
DoRA-64   & 0\%            & 73.50                              & 77.36                                   \\
\textbf{PAT-D}    & 25\%           &  72.98                             & 77.02                                   \\ \bottomrule
\end{tabular}}
}
\caption{Downstream task performance of LoRA, DoRA, and PAT. PAT-L and PAT-D denote our PAT with LoRA and DoRA fine-tuning, respectively.}
\label{tab:dora}
\end{table}
\section{Conclusion}
We propose Pruning-Aware Tuning (PAT), a novel structured pruning approach for Large Language Models (LLMs). PAT prunes the hidden dimensions during the fine-tuning, while preserving the linguistic capabilities. We develop a trainable mask to adaptively set channel values to zero, and efficient Hybrid Sparsification Modules to enable pruning of all linear layers accordingly. The efficiency design reduces the training overhead of PAT to levels comparable to traditional LoRA fine-tuning. Additionally, we propose the Identity Loss to enhance the training robustness by decoupling the rotation and scaling properties of the HSMs.
In the zero-shot evaluation, our 30\%-PAT Llama2 7B and 13B models maintains 98\% performance of those achieved from the LoRA fine-tuning.
\bigskip
\section{Acknowledgments}
This work was supported in part by the Strategic Industries and Key Technologies Project of Jiangsu Province under Grant BE2023020-3. 
\appendix
\section{Appendix}
This section serves as the appendix to the main paper.
\subsection{Detailed Main Results}
We evaluate LoRA~\cite{Hu2021LoRALA}, LLM-Pruner~\cite{Ma2023LLMPrunerOT}, SliceGPT~\cite{ashkboos2024slicegpt} and our PAT on 14 tasks. Results are shown in \cref{tab:appendix-llama2-7b,tab:appendix-llama2-13b,tab:appendix-yi-1.5-34b,tab:appendix-gemma7b,tab:appendix-gemma2b}.
\begin{table}[h]
\caption{The table presents the rank values for the traditional LoRA and our PAT. $\mathbf{R}_{LoRA}$ denotes the rank value in LoRA, and $\mathbf{R}_{HIO}$ denotes the rank value in our PAT. The bold text indicates the settings used in the results of the main paper. Traditional LoRA lacks HIO components, hence their $\mathbf{R}_{HIO}$ is denoted as ``-''. The settings for Llama2 7B are discussed in the ``Efficiency and Accuracy Trade-off'' section of the main paper.}
\label{tab:hio-setting}
\begin{tabular}{ccccc}
\hline
\textbf{Model}                                            & \textbf{Method} & $\mathbf{R}_{LoRA}$ & $\mathbf{R}_{HIO}$ & \textbf{\%Param.} \\ \hline
\multicolumn{1}{c|}{\multirow{2}{*}{\textbf{Gemma 2B}}}   & LoRA            & 64                  & -                  & 3.04              \\
\multicolumn{1}{c|}{}                                     & \textbf{PAT}    & \textbf{20}         & \textbf{300}       & \textbf{2.72}     \\ \hline
\multicolumn{1}{c|}{\multirow{2}{*}{\textbf{Gemma 7B}}}   & LoRA            & 64                  & -                  & 2.29              \\
\multicolumn{1}{c|}{}                                     & \textbf{PAT}    & \textbf{20}         & \textbf{300}       & \textbf{1.93}     \\ \hline
\multicolumn{1}{c|}{\multirow{10}{*}{\textbf{Llama2 7B}}} & LoRA            & 8                   & -                  & 0.27              \\
\multicolumn{1}{c|}{}                                     & LoRA            & 64                  & -                  & 2.32              \\
\multicolumn{1}{c|}{}                                     & LoRA            & 128                 & -                  & 4.53              \\
\multicolumn{1}{c|}{}                                     & LoRA            & 256                 & -                  & 8.67              \\
\multicolumn{1}{c|}{}                                     & LoRA            & 512                 & -                  & 15.89             \\
\multicolumn{1}{c|}{}                                     & PAT             & 8                   & 8                  & 0.36              \\
\multicolumn{1}{c|}{}                                     & \textbf{PAT}    & \textbf{20}         & \textbf{200}       & \textbf{2.27}     \\
\multicolumn{1}{c|}{}                                     & PAT             & 16                  & 512                & 4.43              \\
\multicolumn{1}{c|}{}                                     & PAT             & 32                  & 1024               & 8.49              \\
\multicolumn{1}{c|}{}                                     & PAT             & 8                   & Dense              & 14.15             \\ \hline
\multicolumn{1}{c|}{\multirow{2}{*}{\textbf{Llama2 13B}}} & LoRA            & 64                  & -                  & 1.89              \\
\multicolumn{1}{c|}{}                                     & \textbf{PAT}    & \textbf{20}         & \textbf{200}       & \textbf{1.84}     \\ \hline
\multicolumn{1}{c|}{\multirow{2}{*}{\textbf{Yi-1.5 34B}}} & LoRA            & 64                  & -                  & 1.41              \\
\multicolumn{1}{c|}{}                                     & \textbf{PAT}    & \textbf{20}         & \textbf{200}       & \textbf{1.44}     \\ \hline
\end{tabular}
\end{table}

\begin{table*}
\centering
\setlength{\extrarowheight}{0pt}
\addtolength{\extrarowheight}{\aboverulesep}
\addtolength{\extrarowheight}{\belowrulesep}
\setlength{\aboverulesep}{0pt}
\setlength{\belowrulesep}{0pt}
\caption{Zero-shot evaluation of \textbf{Llama2 7B} on 14 public datasets. \textbf{FT} denotes Fine-Tuning, \textbf{P2FT} denotes Pruning before Fine-Tuning, \textbf{FT2P} denotes Fine-Tuning before Pruning, and \textbf{PAT} denotes Pruning-Aware Tuning.}
\label{tab:appendix-llama2-7b}
\resizebox{\textwidth}{!}{
\begin{tabular}{ccc|cccccccccccccc|c} 
\toprule
\textbf{Ratio}                  & \textbf{Method}                               & \textbf{Mode}                                & \textbf{ARC-C}                                 & \textbf{ARC-E}                                 & \textbf{BOOLQ}                                 & \textbf{COPA}                                  & \textbf{HS}                                    & \textbf{MMLU}                                  & \textbf{MULTIRC}                               & \textbf{OBQA}                                  & \textbf{PIQA}                                  & \textbf{RTE}                                   & \textbf{SIQA}                                  & \textbf{WIC}                                   & \textbf{WG}                                    & \textbf{WSC}                                   & \textbf{AVG}                                    \\ 
\hline
\textbf{0\%}                    & LoRA-64                                       & FT                                           & 50.17                                          & 72.49                                          & 59.57                                          & 72.00                                          & 47.51                                          & 42.93                                          & 60.09                                          & 73.20                                          & 64.80                                          & 52.35                                          & 58.50                                          & 51.10                                          & 52.57                                          & 65.39                                          & 58.76                                           \\ 
\hline
\multirow{5}{*}{\textbf{20\% }} & LLM-Pruner                                    & P2FT                                         & 51.19                                          & 71.78                                          & 64.50                                          & 93.00                                          & 34.41                                          & 39.79                                          & 60.89                                          & 68.80                                          & 65.78                                          & 46.93                                          & 59.88                                          & 47.34                                          & 52.64                                          & 62.50                                          & 58.53                                           \\
                                & LLM-Pruner                                    & FT2P                                         & 21.36                                          & 28.04                                          & 61.71                                          & 65.00                                          & 25.07                                          & 23.01                                          & 43.03                                          & 27.20                                          & 52.94                                          & 53.07                                          & 32.80                                          & 49.69                                          & 50.12                                          & 36.54                                          & 40.68                                           \\
                                & SliceGPT                                      & P2FT                                         & 51.19                                          & 65.43                                          & 65.05                                          & 78.00                                          & 34.76                                          & 37.96                                          & 60.07                                          & 67.00                                          & 65.56                                          & 64.26                                          & 58.24                                          & 52.04                                          & 52.09                                          & 57.69                                          & 57.81                                           \\
                                & SliceGPT                                      & FT2P                                         & 29.83                                          & 48.68                                          & 67.37                                          & 74.00                                          & 26.10                                          & 26.82                                          & 46.00                                          & 49.60                                          & 57.29                                          & 55.60                                          & 44.37                                          & 50.00                                          & 49.80                                          & 43.27                                          & 47.77                                           \\
                                & {\cellcolor[rgb]{1,0.804,0.271}}\textbf{Ours} & {\cellcolor[rgb]{1,0.804,0.271}}\textbf{PAT} & {\cellcolor[rgb]{1,0.804,0.271}}\textbf{61.70} & {\cellcolor[rgb]{1,0.804,0.271}}\textbf{71.96} & {\cellcolor[rgb]{1,0.804,0.271}}\textbf{67.22} & {\cellcolor[rgb]{1,0.804,0.271}}\textbf{78.00} & {\cellcolor[rgb]{1,0.804,0.271}}\textbf{49.71} & {\cellcolor[rgb]{1,0.804,0.271}}\textbf{42.58} & {\cellcolor[rgb]{1,0.804,0.271}}\textbf{60.56} & {\cellcolor[rgb]{1,0.804,0.271}}\textbf{73.40} & {\cellcolor[rgb]{1,0.804,0.271}}\textbf{69.48} & {\cellcolor[rgb]{1,0.804,0.271}}\textbf{52.71} & {\cellcolor[rgb]{1,0.804,0.271}}\textbf{60.24} & {\cellcolor[rgb]{1,0.804,0.271}}\textbf{53.45} & {\cellcolor[rgb]{1,0.804,0.271}}\textbf{52.01} & {\cellcolor[rgb]{1,0.804,0.271}}\textbf{61.54} & {\cellcolor[rgb]{1,0.804,0.271}}\textbf{61.04}  \\ 
\hline
\multirow{5}{*}{\textbf{25\% }} & LLM-Pruner                                    & P2FT                                         & 35.25                                          & 57.14                                          & 48.50                                          & 80.00                                          & 26.76                                          & 32.18                                          & 57.98                                          & 61.20                                          & 64.42                                          & 47.29                                          & 56.50                                          & 50.00                                          & 52.33                                          & 65.38                                          & 52.50                                           \\
                                & LLM-Pruner                                    & FT2P                                         & 21.36                                          & 27.34                                          & 61.74                                          & 55.00                                          & 25.11                                          & 22.95                                          & 43.21                                          & 27.20                                          & 50.22                                          & 52.71                                          & 33.01                                          & 50.00                                          & 49.64                                          & 36.54                                          & 39.72                                           \\
                                & SliceGPT                                      & P2FT                                         & 41.02                                          & 58.55                                          & 48.44                                          & 73.00                                          & 38.91                                          & 32.59                                          & 58.35                                          & 60.40                                          & 64.58                                          & 54.87                                          & 45.65                                          & 54.08                                          & 50.67                                          & 60.58                                          & 52.98                                           \\
                                & SliceGPT                                      & FT2P                                         & 24.07                                          & 31.22                                          & 62.42                                          & 68.00                                          & 25.10                                          & 23.57                                          & 42.95                                          & 29.40                                          & 54.62                                          & 52.71                                          & 37.72                                          & 50.00                                          & 49.33                                          & 36.54                                          & 41.97                                           \\
                                & {\cellcolor[rgb]{1,0.804,0.271}}\textbf{Ours} & {\cellcolor[rgb]{1,0.804,0.271}}\textbf{PAT} & {\cellcolor[rgb]{1,0.804,0.271}}\textbf{54.58} & {\cellcolor[rgb]{1,0.804,0.271}}\textbf{70.90} & {\cellcolor[rgb]{1,0.804,0.271}}\textbf{71.56} & {\cellcolor[rgb]{1,0.804,0.271}}\textbf{82.00} & {\cellcolor[rgb]{1,0.804,0.271}}\textbf{46.04} & {\cellcolor[rgb]{1,0.804,0.271}}\textbf{39.55} & {\cellcolor[rgb]{1,0.804,0.271}}\textbf{63.88} & {\cellcolor[rgb]{1,0.804,0.271}}\textbf{72.40} & {\cellcolor[rgb]{1,0.804,0.271}}\textbf{69.26} & {\cellcolor[rgb]{1,0.804,0.271}}\textbf{67.15} & {\cellcolor[rgb]{1,0.804,0.271}}\textbf{60.90} & {\cellcolor[rgb]{1,0.804,0.271}}\textbf{51.57} & {\cellcolor[rgb]{1,0.804,0.271}}\textbf{51.14} & {\cellcolor[rgb]{1,0.804,0.271}}\textbf{39.42} & {\cellcolor[rgb]{1,0.804,0.271}}\textbf{60.02}  \\ 
\hline
\multirow{5}{*}{\textbf{30\% }} & LLM-Pruner                                    & P2FT                                         & 35.59                                          & 49.82                                          & 63.52                                          & 75.50                                          & 37.04                                          & 32.77                                          & 52.71                                          & 50.10                                          & 60.88                                          & 51.44                                          & 44.91                                          & 52.51                                          & 51.07                                          & 50.48                                          & 50.60                                           \\
                                & LLM-Pruner                                    & FT2P                                         & 21.36                                          & 27.87                                          & 61.99                                          & 55.00                                          & 25.10                                          & 22.95                                          & 42.80                                          & 27.80                                          & 49.73                                          & 52.71                                          & 32.80                                          & 50.00                                          & 49.64                                          & 36.54                                          & 39.73                                           \\
                                & SliceGPT                                      & P2FT                                         & 40.00                                          & 49.65                                          & 64.47                                          & 67.00                                          & 37.82                                          & 32.79                                          & 52.92                                          & 50.20                                          & 58.87                                          & 58.66                                          & 47.54                                          & 53.84                                          & 50.95                                          & 48.08                                          & 50.91                                           \\
                                & SliceGPT                                      & FT2P                                         & 21.36                                          & 28.04                                          & 62.17                                          & 58.00                                          & 25.05                                          & 22.97                                          & 42.82                                          & 27.80                                          & 50.54                                          & 52.71                                          & 34.44                                          & 50.00                                          & 49.57                                          & 36.54                                          & 40.14                                           \\
                                & {\cellcolor[rgb]{1,0.804,0.271}}\textbf{Ours} & {\cellcolor[rgb]{1,0.804,0.271}}\textbf{PAT} & {\cellcolor[rgb]{1,0.804,0.271}}\textbf{51.19} & {\cellcolor[rgb]{1,0.804,0.271}}\textbf{65.43} & {\cellcolor[rgb]{1,0.804,0.271}}\textbf{65.05} & {\cellcolor[rgb]{1,0.804,0.271}}\textbf{78.00} & {\cellcolor[rgb]{1,0.804,0.271}}\textbf{34.76} & {\cellcolor[rgb]{1,0.804,0.271}}\textbf{37.96} & {\cellcolor[rgb]{1,0.804,0.271}}\textbf{60.07} & {\cellcolor[rgb]{1,0.804,0.271}}\textbf{67.00} & {\cellcolor[rgb]{1,0.804,0.271}}\textbf{65.56} & {\cellcolor[rgb]{1,0.804,0.271}}\textbf{64.26} & {\cellcolor[rgb]{1,0.804,0.271}}\textbf{58.24} & {\cellcolor[rgb]{1,0.804,0.271}}\textbf{52.04} & {\cellcolor[rgb]{1,0.804,0.271}}\textbf{52.09} & {\cellcolor[rgb]{1,0.804,0.271}}\textbf{57.69} & {\cellcolor[rgb]{1,0.804,0.271}}\textbf{57.81}  \\
\bottomrule
\end{tabular}

}
\end{table*}

\begin{table*}
\centering
\setlength{\extrarowheight}{0pt}
\addtolength{\extrarowheight}{\aboverulesep}
\addtolength{\extrarowheight}{\belowrulesep}
\setlength{\aboverulesep}{0pt}
\setlength{\belowrulesep}{0pt}
\caption{Zero-shot evaluation of \textbf{Llama2 13B} on 14 public datasets. \textbf{FT} denotes Fine-Tuning, \textbf{P2FT} denotes Pruning before Fine-Tuning, \textbf{FT2P} denotes Fine-Tuning before Pruning, and \textbf{PAT} denotes Pruning-Aware Tuning.}
\label{tab:appendix-llama2-13b}
\resizebox{\textwidth}{!}{
\begin{tabular}{ccc|cccccccccccccc|c} 
\toprule
\textbf{Ratio}                  & \textbf{Method}                               & \textbf{Mode}                                & \textbf{ARC-C}                                 & \textbf{ARC-E}                                 & \textbf{BOOLQ}                                 & \textbf{COPA}                                  & \textbf{HS}                                    & \textbf{MMLU}                                  & \textbf{MULTIRC}                               & \textbf{OBQA}                                  & \textbf{PIQA}                                  & \textbf{RTE}                                   & \textbf{SIQA}                                  & \textbf{WIC}                                   & \textbf{WG}                                    & \textbf{WSC}                                   & \textbf{AVG}                                    \\ 
\hline
\textbf{0\%}                    & LoRA-64                                       & FT                                           & 66.44                                          & 81.31                                          & 70.70                                          & 92.00                                          & 61.22                                          & 50.39                                          & 68.50                                          & 81.00                                          & 76.99                                          & 47.29                                          & 66.68                                          & 50.00                                          & 56.43                                          & 65.39                                          & 66.74                                           \\ 
\hline
\multirow{5}{*}{\textbf{20\% }} & LLM-Pruner                                    & P2FT                                         & 59.66                                          & 77.60                                          & 54.07                                          & 95.00                                          & 62.18                                          & 45.53                                          & 66.30                                          & 76.20                                          & 75.52                                          & 70.40                                          & 62.79                                          & 50.47                                          & 53.83                                          & 64.42                                          & 65.28                                           \\
                                & LLM-Pruner                                    & FT2P                                         & 27.12                                          & 27.34                                          & 62.20                                          & 55.00                                          & 27.89                                          & 26.84                                          & 42.80                                          & 32.20                                          & 49.51                                          & 52.71                                          & 40.33                                          & 50.00                                          & 49.57                                          & 36.54                                          & 41.43                                           \\
                                & SliceGPT                                      & P2FT                                         & 61.19                                          & 77.25                                          & 77.46                                          & 88.50                                          & 57.11                                          & 43.22                                          & 72.03                                          & 78.50                                          & 71.60                                          & 70.76                                          & 63.84                                          & 53.68                                          & 56.00                                          & 50.96                                          & 65.86                                           \\
                                & SliceGPT                                      & FT2P                                         & 32.88                                          & 42.68                                          & 78.81                                          & 50.00                                          & 38.99                                          & 31.57                                          & 51.40                                          & 47.60                                          & 65.18                                          & 57.40                                          & 50.36                                          & 55.02                                          & 50.75                                          & 56.73                                          & 50.67                                           \\
                                & {\cellcolor[rgb]{1,0.804,0.271}}\textbf{Ours} & {\cellcolor[rgb]{1,0.804,0.271}}\textbf{PAT} & {\cellcolor[rgb]{1,0.804,0.271}}\textbf{65.76} & {\cellcolor[rgb]{1,0.804,0.271}}\textbf{80.95} & {\cellcolor[rgb]{1,0.804,0.271}}\textbf{74.56} & {\cellcolor[rgb]{1,0.804,0.271}}\textbf{89.00} & {\cellcolor[rgb]{1,0.804,0.271}}\textbf{62.88} & {\cellcolor[rgb]{1,0.804,0.271}}\textbf{47.95} & {\cellcolor[rgb]{1,0.804,0.271}}\textbf{75.60} & {\cellcolor[rgb]{1,0.804,0.271}}\textbf{81.60} & {\cellcolor[rgb]{1,0.804,0.271}}\textbf{72.74} & {\cellcolor[rgb]{1,0.804,0.271}}\textbf{79.42} & {\cellcolor[rgb]{1,0.804,0.271}}\textbf{65.81} & {\cellcolor[rgb]{1,0.804,0.271}}\textbf{54.39} & {\cellcolor[rgb]{1,0.804,0.271}}\textbf{55.17} & {\cellcolor[rgb]{1,0.804,0.271}}\textbf{65.39} & {\cellcolor[rgb]{1,0.804,0.271}}\textbf{69.37}  \\ 
\hline
\multirow{5}{*}{\textbf{25\% }} & LLM-Pruner                                    & P2FT                                         & 60.00                                          & 70.72                                          & 40.64                                          & 71.00                                          & 57.98                                          & 45.51                                          & 59.78                                          & 74.40                                          & 71.76                                          & 47.29                                          & 56.65                                          & 50.31                                          & 51.38                                          & 63.46                                          & 58.64                                           \\
                                & LLM-Pruner                                    & FT2P                                         & 22.37                                          & 28.57                                          & 62.11                                          & 55.00                                          & 25.02                                          & 23.37                                          & 42.82                                          & 27.00                                          & 49.46                                          & 52.71                                          & 32.91                                          & 50.00                                          & 49.57                                          & 36.54                                          & 39.82                                           \\
                                & SliceGPT                                      & P2FT                                         & 62.20                                          & 71.96                                          & 59.72                                          & 70.50                                          & 56.80                                          & 43.78                                          & 64.32                                          & 72.40                                          & 67.52                                          & 61.19                                          & 58.80                                          & 52.90                                          & 53.75                                          & 53.85                                          & 60.69                                           \\
                                & SliceGPT                                      & FT2P                                         & 33.56                                          & 38.45                                          & 67.95                                          & 77.00                                          & 33.06                                          & 24.79                                          & 42.88                                          & 43.60                                          & 54.46                                          & 52.71                                          & 47.59                                          & 50.00                                          & 50.91                                          & 37.50                                          & 46.75                                           \\
                                & {\cellcolor[rgb]{1,0.804,0.271}}\textbf{Ours} & {\cellcolor[rgb]{1,0.804,0.271}}\textbf{PAT} & {\cellcolor[rgb]{1,0.804,0.271}}\textbf{66.10} & {\cellcolor[rgb]{1,0.804,0.271}}\textbf{77.95} & {\cellcolor[rgb]{1,0.804,0.271}}\textbf{81.01} & {\cellcolor[rgb]{1,0.804,0.271}}\textbf{84.00} & {\cellcolor[rgb]{1,0.804,0.271}}\textbf{60.73} & {\cellcolor[rgb]{1,0.804,0.271}}\textbf{43.57} & {\cellcolor[rgb]{1,0.804,0.271}}\textbf{70.38} & {\cellcolor[rgb]{1,0.804,0.271}}\textbf{79.40} & {\cellcolor[rgb]{1,0.804,0.271}}\textbf{72.25} & {\cellcolor[rgb]{1,0.804,0.271}}\textbf{75.09} & {\cellcolor[rgb]{1,0.804,0.271}}\textbf{64.79} & {\cellcolor[rgb]{1,0.804,0.271}}\textbf{55.80} & {\cellcolor[rgb]{1,0.804,0.271}}\textbf{56.83} & {\cellcolor[rgb]{1,0.804,0.271}}\textbf{44.23} & {\cellcolor[rgb]{1,0.804,0.271}}\textbf{66.58}  \\ 
\hline
\multirow{5}{*}{\textbf{30\% }} & LLM-Pruner                                    & P2FT                                         & 40.85                                          & 51.06                                          & 65.14                                          & 63.50                                          & 42.89                                          & 35.87                                          & 54.92                                          & 53.80                                          & 56.28                                          & 50.36                                          & 50.72                                          & 50.00                                          & 51.10                                          & 51.44                                          & 51.28                                           \\
                                & LLM-Pruner                                    & FT2P                                         & 21.36                                          & 27.87                                          & 62.14                                          & 55.00                                          & 25.06                                          & 22.91                                          & 42.80                                          & 27.40                                          & 49.51                                          & 52.71                                          & 32.91                                          & 50.00                                          & 49.57                                          & 36.54                                          & 39.70                                           \\
                                & SliceGPT                                      & P2FT                                         & 51.19                                          & 58.47                                          & 58.88                                          & 70.00                                          & 47.31                                          & 39.55                                          & 64.99                                          & 62.10                                          & 65.53                                          & 47.65                                          & 51.89                                          & 50.00                                          & 53.63                                          & 64.42                                          & 56.12                                           \\
                                & SliceGPT                                      & FT2P                                         & 32.88                                          & 43.21                                          & 68.44                                          & 63.00                                          & 30.46                                          & 24.48                                          & 42.90                                          & 47.00                                          & 59.85                                          & 53.07                                          & 44.58                                          & 50.00                                          & 50.28                                          & 36.54                                          & 46.19                                           \\
                                & {\cellcolor[rgb]{1,0.804,0.271}}\textbf{Ours} & {\cellcolor[rgb]{1,0.804,0.271}}\textbf{PAT} & {\cellcolor[rgb]{1,0.804,0.271}}\textbf{56.27} & {\cellcolor[rgb]{1,0.804,0.271}}\textbf{76.54} & {\cellcolor[rgb]{1,0.804,0.271}}\textbf{73.91} & {\cellcolor[rgb]{1,0.804,0.271}}\textbf{93.00} & {\cellcolor[rgb]{1,0.804,0.271}}\textbf{53.50} & {\cellcolor[rgb]{1,0.804,0.271}}\textbf{42.87} & {\cellcolor[rgb]{1,0.804,0.271}}\textbf{73.68} & {\cellcolor[rgb]{1,0.804,0.271}}\textbf{77.60} & {\cellcolor[rgb]{1,0.804,0.271}}\textbf{70.95} & {\cellcolor[rgb]{1,0.804,0.271}}\textbf{66.43} & {\cellcolor[rgb]{1,0.804,0.271}}\textbf{62.90} & {\cellcolor[rgb]{1,0.804,0.271}}\textbf{51.57} & {\cellcolor[rgb]{1,0.804,0.271}}\textbf{55.17} & {\cellcolor[rgb]{1,0.804,0.271}}\textbf{57.69} & {\cellcolor[rgb]{1,0.804,0.271}}\textbf{65.15}  \\
\bottomrule
\end{tabular}
}
\end{table*}

\begin{table*}
\centering
\setlength{\extrarowheight}{0pt}
\addtolength{\extrarowheight}{\aboverulesep}
\addtolength{\extrarowheight}{\belowrulesep}
\setlength{\aboverulesep}{0pt}
\setlength{\belowrulesep}{0pt}
\caption{Zero-shot evaluation of \textbf{Yi-1.5 34B} on 14 public datasets. \textbf{FT} denotes Fine-Tuning, \textbf{P2FT} denotes Pruning before Fine-Tuning, \textbf{FT2P} denotes Fine-Tuning before Pruning, and \textbf{PAT} denotes Pruning-Aware Tuning.}
\label{tab:appendix-yi-1.5-34b}
\resizebox{\textwidth}{!}{
\begin{tabular}{ccc|cccccccccccccc|c} 
\toprule
\textbf{Ratio}                  & \textbf{Method}                               & \textbf{Mode}                                & \textbf{ARC-C}                                 & \textbf{ARC-E}                                 & \textbf{BOOLQ}                                 & \textbf{COPA}                                  & \textbf{HS}                                    & \textbf{MMLU}                                  & \textbf{MULTIRC}                               & \textbf{OBQA}                                  & \textbf{PIQA}                                  & \textbf{RTE}                                   & \textbf{SIQA}                                  & \textbf{WIC}                                   & \textbf{WG}                                    & \textbf{WSC}                                   & \textbf{AVG}                                    \\ 
\hline
\textbf{0\%}                    & LoRA-64                                       & FT                                           & 91.19                                          & 96.65                                          & 89.11                                          & 99.00                                          & 82.54                                          & 68.97                                          & 84.32                                          & 92.40                                          & 87.16                                          & 75.81                                          & 76.15                                          & 62.54                                          & 71.43                                          & 59.62                                          & 81.21                                           \\ 
\hline
\multirow{5}{*}{\textbf{20\% }} & LLM-Pruner                                    & P2FT                                         & 79.15                                          & 92.15                                          & 86.35                                          & 95.50                                          & 72.74                                          & 57.44                                          & 68.88                                          & 87.50                                          & 80.85                                          & 68.23                                          & 71.49                                          & 59.80                                          & 63.06                                          & 50.96                                          & 73.86                                           \\
                                & LLM-Pruner                                    & FT2P                                         & 41.19                                          & 64.20                                          & 66.44                                          & 82.00                                          & 39.19                                          & 36.09                                          & 47.99                                          & 68.10                                          & 64.15                                          & 54.87                                          & 50.41                                          & 51.65                                          & 51.58                                          & 36.54                                          & 53.88                                           \\
                                & SliceGPT                                      & P2FT                                         & 85.08                                          & 93.74                                          & 87.40                                          & 96.50                                          & 79.49                                          & 62.42                                          & 70.85                                          & 88.90                                          & 86.32                                          & 68.77                                          & 74.21                                          & 61.21                                          & 65.19                                          & 55.29                                          & 76.81                                           \\
                                & SliceGPT                                      & FT2P                                         & 68.31                                          & 88.18                                          & 83.88                                          & 93.00                                          & 65.26                                          & 46.93                                          & 56.11                                          & 83.90                                          & 75.41                                          & 60.83                                          & 67.78                                          & 57.13                                          & 54.54                                          & 45.19                                          & 67.60                                           \\
                                & {\cellcolor[rgb]{1,0.804,0.271}}\textbf{Ours} & {\cellcolor[rgb]{1,0.804,0.271}}\textbf{PAT} & {\cellcolor[rgb]{1,0.804,0.271}}\textbf{91.53} & {\cellcolor[rgb]{1,0.804,0.271}}\textbf{96.83} & {\cellcolor[rgb]{1,0.804,0.271}}\textbf{88.13} & {\cellcolor[rgb]{1,0.804,0.271}}\textbf{99.00} & {\cellcolor[rgb]{1,0.804,0.271}}\textbf{82.66} & {\cellcolor[rgb]{1,0.804,0.271}}\textbf{65.05} & {\cellcolor[rgb]{1,0.804,0.271}}\textbf{84.09} & {\cellcolor[rgb]{1,0.804,0.271}}\textbf{92.40} & {\cellcolor[rgb]{1,0.804,0.271}}\textbf{87.81} & {\cellcolor[rgb]{1,0.804,0.271}}\textbf{78.52} & {\cellcolor[rgb]{1,0.804,0.271}}\textbf{74.41} & {\cellcolor[rgb]{1,0.804,0.271}}\textbf{61.91} & {\cellcolor[rgb]{1,0.804,0.271}}\textbf{71.11} & {\cellcolor[rgb]{1,0.804,0.271}}\textbf{61.54} & {\cellcolor[rgb]{1,0.804,0.271}}\textbf{81.02}  \\ 
\hline
\multirow{5}{*}{\textbf{25\% }} & LLM-Pruner                                    & P2FT                                         & 70.34                                          & 86.86                                          & 83.64                                          & 94.00                                          & 67.59                                          & 49.66                                          & 63.43                                          & 83.90                                          & 79.54                                          & 66.79                                          & 69.50                                          & 54.62                                          & 58.13                                          & 53.37                                          & 70.10                                           \\
                                & LLM-Pruner                                    & FT2P                                         & 39.32                                          & 54.94                                          & 58.33                                          & 73.00                                          & 38.30                                          & 32.67                                          & 51.18                                          & 56.00                                          & 60.69                                          & 53.97                                          & 44.73                                          & 52.04                                          & 50.67                                          & 48.56                                          & 51.03                                           \\
                                & SliceGPT                                      & P2FT                                         & 80.00                                          & 91.27                                          & 86.56                                          & 94.00                                          & 75.41                                          & 55.68                                          & 62.13                                          & 87.10                                          & 84.79                                          & 67.51                                          & 72.77                                          & 60.50                                          & 61.33                                          & 55.29                                          & 73.88                                           \\
                                & SliceGPT                                      & FT2P                                         & 57.80                                          & 62.08                                          & 63.55                                          & 72.00                                          & 53.59                                          & 46.84                                          & 70.75                                          & 58.20                                          & 68.83                                          & 61.55                                          & 54.89                                          & 56.35                                          & 60.93                                          & 61.54                                          & 60.63                                           \\
                                & {\cellcolor[rgb]{1,0.804,0.271}}\textbf{Ours} & {\cellcolor[rgb]{1,0.804,0.271}}\textbf{PAT} & {\cellcolor[rgb]{1,0.804,0.271}}\textbf{83.39} & {\cellcolor[rgb]{1,0.804,0.271}}\textbf{91.53} & {\cellcolor[rgb]{1,0.804,0.271}}\textbf{87.16} & {\cellcolor[rgb]{1,0.804,0.271}}\textbf{99.00} & {\cellcolor[rgb]{1,0.804,0.271}}\textbf{78.75} & {\cellcolor[rgb]{1,0.804,0.271}}\textbf{61.14} & {\cellcolor[rgb]{1,0.804,0.271}}\textbf{83.85} & {\cellcolor[rgb]{1,0.804,0.271}}\textbf{89.60} & {\cellcolor[rgb]{1,0.804,0.271}}\textbf{83.90} & {\cellcolor[rgb]{1,0.804,0.271}}\textbf{81.23} & {\cellcolor[rgb]{1,0.804,0.271}}\textbf{74.41} & {\cellcolor[rgb]{1,0.804,0.271}}\textbf{62.70} & {\cellcolor[rgb]{1,0.804,0.271}}\textbf{67.32} & {\cellcolor[rgb]{1,0.804,0.271}}\textbf{60.58} & {\cellcolor[rgb]{1,0.804,0.271}}\textbf{78.90}  \\ 
\hline
\multirow{5}{*}{\textbf{30\% }} & LLM-Pruner                                    & P2FT                                         & 70.85                                          & 78.75                                          & 62.80                                          & 78.00                                          & 65.81                                          & 52.56                                          & 71.05                                          & 77.50                                          & 73.34                                          & 64.26                                          & 63.61                                          & 56.35                                          & 59.00                                          & 62.02                                          & 66.85                                           \\
                                & LLM-Pruner                                    & FT2P                                         & 30.85                                          & 41.31                                          & 60.10                                          & 63.00                                          & 31.59                                          & 28.22                                          & 46.99                                          & 41.80                                          & 55.10                                          & 53.34                                          & 39.00                                          & 50.86                                          & 50.28                                          & 42.55                                          & 45.36                                           \\
                                & SliceGPT                                      & P2FT                                         & 73.90                                          & 86.60                                          & 80.69                                          & 91.00                                          & 69.84                                          & 52.14                                          & 68.38                                          & 84.60                                          & 78.84                                          & 72.74                                          & 69.45                                          & 57.60                                          & 58.49                                          & 61.06                                          & 71.81                                           \\
                                & SliceGPT                                      & FT2P                                         & 50.85                                          & 60.03                                          & 61.45                                          & 70.50                                          & 48.70                                          & 40.39                                          & 59.02                                          & 59.65                                          & 64.22                                          & 58.80                                          & 51.30                                          & 53.61                                          & 54.64                                          & 52.28                                          & 56.10                                           \\
                                & {\cellcolor[rgb]{1,0.804,0.271}}\textbf{Ours} & {\cellcolor[rgb]{1,0.804,0.271}}\textbf{PAT} & {\cellcolor[rgb]{1,0.804,0.271}}\textbf{86.31} & {\cellcolor[rgb]{1,0.804,0.271}}\textbf{92.15} & {\cellcolor[rgb]{1,0.804,0.271}}\textbf{87.69} & {\cellcolor[rgb]{1,0.804,0.271}}\textbf{95.50} & {\cellcolor[rgb]{1,0.804,0.271}}\textbf{79.17} & {\cellcolor[rgb]{1,0.804,0.271}}\textbf{63.15} & {\cellcolor[rgb]{1,0.804,0.271}}\textbf{80.29} & {\cellcolor[rgb]{1,0.804,0.271}}\textbf{90.48} & {\cellcolor[rgb]{1,0.804,0.271}}\textbf{84.93} & {\cellcolor[rgb]{1,0.804,0.271}}\textbf{75.95} & {\cellcolor[rgb]{1,0.804,0.271}}\textbf{71.49} & {\cellcolor[rgb]{1,0.804,0.271}}\textbf{61.74} & {\cellcolor[rgb]{1,0.804,0.271}}\textbf{63.06} & {\cellcolor[rgb]{1,0.804,0.271}}\textbf{58.17} & {\cellcolor[rgb]{1,0.804,0.271}}\textbf{77.89}  \\
\bottomrule
\end{tabular}
}
\end{table*}

\begin{table*}
\centering
\setlength{\extrarowheight}{0pt}
\addtolength{\extrarowheight}{\aboverulesep}
\addtolength{\extrarowheight}{\belowrulesep}
\setlength{\aboverulesep}{0pt}
\setlength{\belowrulesep}{0pt}
\caption{Zero-shot evaluation of \textbf{Gemma 7B} on 14 public datasets. \textbf{FT} denotes Fine-Tuning, \textbf{P2FT} denotes Pruning before Fine-Tuning, \textbf{FT2P} denotes Fine-Tuning before Pruning, and \textbf{PAT} denotes Pruning-Aware Tuning.}
\label{tab:appendix-gemma7b}
\resizebox{\textwidth}{!}{
\begin{tabular}{ccc|cccccccccccccc|c} 
\toprule
\textbf{Ratio}                  & \textbf{Method}                               & \textbf{Mode}                                & \textbf{ARC-C}                        & \textbf{ARC-E}                        & \textbf{BOOLQ}                        & \textbf{COPA}                         & \textbf{HS}                           & \textbf{MMLU}                         & \textbf{MULTIRC}                      & \textbf{OBQA}                         & \textbf{PIQA}                         & \textbf{RTE}                          & \textbf{SIQA}                         & \textbf{WIC}                          & \textbf{WG}                           & \textbf{WSC}                          & \textbf{AVG}                           \\ 
\hline
\textbf{0\%}                    & LoRA-64                                       & FT                                           & 80.00                                 & 89.77                                 & 85.05                                 & 93.00                                 & 74.57                                 & 56.82                                 & 54.41                                 & 85.60                                 & 84.06                                 & 60.29                                 & 68.94                                 & 57.68                                 & 57.22                                 & 54.81                                 & 71.59                                  \\ 
\hline
\multirow{5}{*}{\textbf{20\% }} & LLM-Pruner                                    & P2FT                                         & 64.83                                 & 81.04                                 & 76.01                                 & 83.25                                 & 62.62                                 & 46.36                                 & 60.54                                 & 80.25                                 & 72.58                                 & 58.84                                 & 64.29                                 & 56.31                                 & 55.31                                 & 54.09                                 & 65.45                                  \\
                                & LLM-Pruner                                    & FT2P                                         & 47.49                                 & 59.16                                 & 66.66                                 & 70.08                                 & 46.81                                 & 37.00                                 & 55.38                                 & 57.85                                 & 63.29                                 & 55.70                                 & 51.62                                 & 53.36                                 & 53.12                                 & 50.72                                 & 54.87                                  \\
                                & SliceGPT                                      & P2FT                                         & 64.92                                 & 85.54                                 & 82.28                                 & 92.50                                 & 62.05                                 & 45.10                                 & 58.60                                 & 81.90                                 & 74.48                                 & 62.64                                 & 66.68                                 & 53.06                                 & 55.05                                 & 47.60                                 & 66.60                                  \\
                                & SliceGPT                                      & FT2P                                         & 49.15                                 & 62.20                                 & 70.66                                 & 72.67                                 & 48.20                                 & 37.88                                 & 54.57                                 & 61.60                                 & 64.29                                 & 57.34                                 & 53.27                                 & 52.72                                 & 53.72                                 & 48.08                                 & 56.17                                  \\
                                & {\cellcolor[rgb]{1,0.804,0.271}}\textbf{Ours} & {\cellcolor[rgb]{1,0.804,0.271}}\textbf{PAT} & {\cellcolor[rgb]{1,0.804,0.271}}69.49 & {\cellcolor[rgb]{1,0.804,0.271}}88.71 & {\cellcolor[rgb]{1,0.804,0.271}}84.19 & {\cellcolor[rgb]{1,0.804,0.271}}94.00 & {\cellcolor[rgb]{1,0.804,0.271}}67.58 & {\cellcolor[rgb]{1,0.804,0.271}}47.95 & {\cellcolor[rgb]{1,0.804,0.271}}58.77 & {\cellcolor[rgb]{1,0.804,0.271}}85.20 & {\cellcolor[rgb]{1,0.804,0.271}}76.28 & {\cellcolor[rgb]{1,0.804,0.271}}61.01 & {\cellcolor[rgb]{1,0.804,0.271}}68.73 & {\cellcolor[rgb]{1,0.804,0.271}}57.21 & {\cellcolor[rgb]{1,0.804,0.271}}54.38 & {\cellcolor[rgb]{1,0.804,0.271}}48.08 & {\cellcolor[rgb]{1,0.804,0.271}}68.68  \\ 
\hline
\multirow{5}{*}{\textbf{25\% }} & LLM-Pruner                                    & P2FT                                         & 55.89                                 & 71.87                                 & 77.41                                 & 83.88                                 & 55.83                                 & 41.05                                 & 53.30                                 & 68.88                                 & 69.41                                 & 62.05                                 & 59.51                                 & 51.70                                 & 54.09                                 & 42.19                                 & 60.50                                  \\
                                & LLM-Pruner                                    & FT2P                                         & 40.56                                 & 50.81                                 & 59.62                                 & 64.64                                 & 39.53                                 & 32.76                                 & 53.62                                 & 47.35                                 & 59.68                                 & 53.50                                 & 46.47                                 & 52.54                                 & 52.02                                 & 51.04                                 & 50.29                                  \\
                                & SliceGPT                                      & P2FT                                         & 60.17                                 & 73.37                                 & 67.83                                 & 72.50                                 & 57.67                                 & 44.77                                 & 62.30                                 & 75.30                                 & 68.88                                 & 56.68                                 & 59.85                                 & 55.41                                 & 56.24                                 & 60.10                                 & 62.22                                  \\
                                & SliceGPT                                      & FT2P                                         & 44.75                                 & 54.94                                 & 59.37                                 & 67.00                                 & 41.50                                 & 33.98                                 & 54.49                                 & 51.90                                 & 61.92                                 & 57.40                                 & 49.87                                 & 50.31                                 & 51.38                                 & 50.96                                 & 52.13                                  \\
                                & {\cellcolor[rgb]{1,0.804,0.271}}\textbf{Ours} & {\cellcolor[rgb]{1,0.804,0.271}}\textbf{PAT} & {\cellcolor[rgb]{1,0.804,0.271}}62.71 & {\cellcolor[rgb]{1,0.804,0.271}}83.42 & {\cellcolor[rgb]{1,0.804,0.271}}80.98 & {\cellcolor[rgb]{1,0.804,0.271}}93.00 & {\cellcolor[rgb]{1,0.804,0.271}}61.16 & {\cellcolor[rgb]{1,0.804,0.271}}44.30 & {\cellcolor[rgb]{1,0.804,0.271}}63.76 & {\cellcolor[rgb]{1,0.804,0.271}}81.20 & {\cellcolor[rgb]{1,0.804,0.271}}74.43 & {\cellcolor[rgb]{1,0.804,0.271}}64.62 & {\cellcolor[rgb]{1,0.804,0.271}}66.53 & {\cellcolor[rgb]{1,0.804,0.271}}49.06 & {\cellcolor[rgb]{1,0.804,0.271}}55.41 & {\cellcolor[rgb]{1,0.804,0.271}}52.89 & {\cellcolor[rgb]{1,0.804,0.271}}66.68  \\ 
\hline
\multirow{5}{*}{\textbf{30\% }} & LLM-Pruner                                    & P2FT                                         & 40.15                                 & 49.69                                 & 57.70                                 & 64.44                                 & 40.23                                 & 32.89                                 & 55.24                                 & 46.44                                 & 59.95                                 & 54.67                                 & 46.56                                 & 50.93                                 & 52.26                                 & 52.82                                 & 50.28                                  \\
                                & LLM-Pruner                                    & FT2P                                         & 25.64                                 & 30.83                                 & 52.38                                 & 51.64                                 & 26.84                                 & 25.38                                 & 50.07                                 & 29.02                                 & 51.33                                 & 50.85                                 & 35.28                                 & 50.08                                 & 50.41                                 & 49.12                                 & 41.35                                  \\
                                & SliceGPT                                      & P2FT                                         & 43.64                                 & 56.61                                 & 72.07                                 & 72.75                                 & 43.47                                 & 34.44                                 & 50.70                                 & 54.75                                 & 62.00                                 & 57.67                                 & 49.97                                 & 51.37                                 & 52.47                                 & 42.07                                 & 53.14                                  \\
                                & SliceGPT                                      & FT2P                                         & 30.14                                 & 37.27                                 & 57.31                                 & 56.92                                 & 30.99                                 & 27.64                                 & 50.23                                 & 35.45                                 & 54.00                                 & 52.56                                 & 38.95                                 & 50.41                                 & 50.93                                 & 47.36                                 & 44.30                                  \\
                                & {\cellcolor[rgb]{1,0.804,0.271}}\textbf{Ours} & {\cellcolor[rgb]{1,0.804,0.271}}\textbf{PAT} & {\cellcolor[rgb]{1,0.804,0.271}}64.75 & {\cellcolor[rgb]{1,0.804,0.271}}82.01 & {\cellcolor[rgb]{1,0.804,0.271}}80.55 & {\cellcolor[rgb]{1,0.804,0.271}}90.00 & {\cellcolor[rgb]{1,0.804,0.271}}59.27 & {\cellcolor[rgb]{1,0.804,0.271}}43.36 & {\cellcolor[rgb]{1,0.804,0.271}}57.10 & {\cellcolor[rgb]{1,0.804,0.271}}79.60 & {\cellcolor[rgb]{1,0.804,0.271}}73.12 & {\cellcolor[rgb]{1,0.804,0.271}}67.87 & {\cellcolor[rgb]{1,0.804,0.271}}65.97 & {\cellcolor[rgb]{1,0.804,0.271}}49.37 & {\cellcolor[rgb]{1,0.804,0.271}}51.78 & {\cellcolor[rgb]{1,0.804,0.271}}39.42 & {\cellcolor[rgb]{1,0.804,0.271}}64.58  \\
\bottomrule
\end{tabular}
}
\end{table*}

\begin{table*}
\centering
\setlength{\extrarowheight}{0pt}
\addtolength{\extrarowheight}{\aboverulesep}
\addtolength{\extrarowheight}{\belowrulesep}
\setlength{\aboverulesep}{0pt}
\setlength{\belowrulesep}{0pt}
\caption{Zero-shot evaluation of \textbf{Gemma 2B} on 14 public datasets. \textbf{FT} denotes Fine-Tuning, \textbf{P2FT} denotes Pruning before Fine-Tuning, \textbf{FT2P} denotes Fine-Tuning before Pruning, and \textbf{PAT} denotes Pruning-Aware Tuning.}
\label{tab:appendix-gemma2b}
\resizebox{\textwidth}{!}{
\begin{tabular}{ccc|cccccccccccccc|c} 
\toprule
\textbf{Ratio}                  & \textbf{Method}                               & \textbf{Mode}                                & \textbf{ARC-C}                        & \textbf{ARC-E}                        & \textbf{BOOLQ}                        & \textbf{COPA}                         & \textbf{HS}                           & \textbf{MMLU}                         & \textbf{MULTIRC}                      & \textbf{OBQA}                         & \textbf{PIQA}                         & \textbf{RTE}                          & \textbf{SIQA}                         & \textbf{WIC}                          & \textbf{WG}                           & \textbf{WSC}                          & \textbf{AVG}                                    \\ 
\hline
\textbf{0\%}                    & LoRA-64                                       & FT                                           & 42.71                                 & 66.49                                 & 63.73                                 & 88.00                                 & 39.19                                 & 38.08                                 & 41.50                                 & 67.40                                 & 64.85                                 & 52.71                                 & 50.31                                 & 49.69                                 & 52.33                                 & 36.54                                 & 53.82                                           \\ 
\hline
\multirow{5}{*}{\textbf{20\% }} & LLM-Pruner                                    & P2FT                                         & 34.58                                 & 46.56                                 & 48.24                                 & 58.50                                 & 34.07                                 & 31.52                                 & 63.22                                 & 56.60                                 & 56.09                                 & 48.92                                 & 41.48                                 & 50.00                                 & 50.99                                 & 63.46                                 & 48.87                                           \\
                                & LLM-Pruner                                    & FT2P                                         & 24.58                                 & 28.54                                 & 55.51                                 & 52.00                                 & 26.24                                 & 24.60                                 & 48.35                                 & 29.03                                 & 50.26                                 & 51.02                                 & 33.40                                 & 50.00                                 & 49.89                                 & 45.51                                 & 40.64                                           \\
                                & SliceGPT                                      & P2FT                                         & 34.92                                 & 45.86                                 & 55.44                                 & 54.00                                 & 34.65                                 & 30.84                                 & 60.81                                 & 49.00                                 & 52.34                                 & 55.24                                 & 44.63                                 & 46.08                                 & 50.59                                 & 60.58                                 & 48.21                                           \\
                                & SliceGPT                                      & FT2P                                         & 26.22                                 & 31.86                                 & 53.83                                 & 57.00                                 & 26.33                                 & 24.82                                 & 47.62                                 & 30.73                                 & 52.38                                 & 50.90                                 & 36.40                                 & 50.00                                 & 50.43                                 & 44.87                                 & 41.67                                           \\
                                & {\cellcolor[rgb]{1,0.804,0.271}}\textbf{Ours} & {\cellcolor[rgb]{1,0.804,0.271}}\textbf{PAT} & {\cellcolor[rgb]{1,0.804,0.271}}39.66 & {\cellcolor[rgb]{1,0.804,0.271}}61.91 & {\cellcolor[rgb]{1,0.804,0.271}}69.14 & {\cellcolor[rgb]{1,0.804,0.271}}76.00 & {\cellcolor[rgb]{1,0.804,0.271}}39.20 & {\cellcolor[rgb]{1,0.804,0.271}}34.09 & {\cellcolor[rgb]{1,0.804,0.271}}54.48 & {\cellcolor[rgb]{1,0.804,0.271}}68.80 & {\cellcolor[rgb]{1,0.804,0.271}}63.44 & {\cellcolor[rgb]{1,0.804,0.271}}57.04 & {\cellcolor[rgb]{1,0.804,0.271}}50.51 & {\cellcolor[rgb]{1,0.804,0.271}}53.61 & {\cellcolor[rgb]{1,0.804,0.271}}50.83 & {\cellcolor[rgb]{1,0.804,0.271}}36.54 & {\cellcolor[rgb]{1,0.804,0.271}}\textbf{53.95}  \\ 
\hline
\multirow{5}{*}{\textbf{25\% }} & LLM-Pruner                                    & P2FT                                         & 27.80                                 & 31.13                                 & 54.40                                 & 51.00                                 & 28.61                                 & 25.84                                 & 52.25                                 & 35.20                                 & 51.31                                 & 50.36                                 & 34.42                                 & 50.00                                 & 50.12                                 & 50.00                                 & 42.32                                           \\
                                & LLM-Pruner                                    & FT2P                                         & 21.36                                 & 27.69                                 & 61.73                                 & 60.00                                 & 25.09                                 & 22.98                                 & 43.12                                 & 27.20                                 & 51.58                                 & 52.89                                 & 32.91                                 & 49.84                                 & 49.88                                 & 36.54                                 & 40.20                                           \\
                                & SliceGPT                                      & P2FT                                         & 29.49                                 & 42.33                                 & 61.56                                 & 71.00                                 & 28.90                                 & 24.38                                 & 42.86                                 & 43.40                                 & 57.18                                 & 52.71                                 & 43.45                                 & 50.00                                 & 51.30                                 & 34.62                                 & 45.23                                           \\
                                & SliceGPT                                      & FT2P                                         & 21.36                                 & 27.87                                 & 62.17                                 & 55.00                                 & 25.05                                 & 22.95                                 & 42.80                                 & 27.60                                 & 49.51                                 & 52.71                                 & 32.91                                 & 50.00                                 & 49.57                                 & 36.54                                 & 39.72                                           \\
                                & {\cellcolor[rgb]{1,0.804,0.271}}\textbf{Ours} & {\cellcolor[rgb]{1,0.804,0.271}}\textbf{PAT} & {\cellcolor[rgb]{1,0.804,0.271}}41.02 & {\cellcolor[rgb]{1,0.804,0.271}}58.55 & {\cellcolor[rgb]{1,0.804,0.271}}48.44 & {\cellcolor[rgb]{1,0.804,0.271}}73.00 & {\cellcolor[rgb]{1,0.804,0.271}}38.91 & {\cellcolor[rgb]{1,0.804,0.271}}32.59 & {\cellcolor[rgb]{1,0.804,0.271}}58.35 & {\cellcolor[rgb]{1,0.804,0.271}}60.40 & {\cellcolor[rgb]{1,0.804,0.271}}64.58 & {\cellcolor[rgb]{1,0.804,0.271}}54.87 & {\cellcolor[rgb]{1,0.804,0.271}}45.65 & {\cellcolor[rgb]{1,0.804,0.271}}54.08 & {\cellcolor[rgb]{1,0.804,0.271}}50.67 & {\cellcolor[rgb]{1,0.804,0.271}}60.58 & {\cellcolor[rgb]{1,0.804,0.271}}\textbf{52.98}  \\ 
\hline
\multirow{5}{*}{\textbf{30\% }} & LLM-Pruner                                    & P2FT                                         & 21.36                                 & 27.87                                 & 62.16                                 & 55.00                                 & 25.05                                 & 22.93                                 & 42.80                                 & 27.50                                 & 49.51                                 & 52.71                                 & 32.91                                 & 50.00                                 & 49.57                                 & 36.54                                 & 39.71                                           \\
                                & LLM-Pruner                                    & FT2P                                         & 21.36                                 & 27.75                                 & 61.81                                 & 58.33                                 & 25.10                                 & 22.97                                 & 43.01                                 & 27.40                                 & 50.96                                 & 52.83                                 & 32.87                                 & 49.90                                 & 49.80                                 & 36.54                                 & 40.05                                           \\
                                & SliceGPT                                      & P2FT                                         & 27.80                                 & 25.40                                 & 37.77                                 & 45.00                                 & 25.06                                 & 27.12                                 & 57.20                                 & 21.20                                 & 50.44                                 & 47.29                                 & 32.86                                 & 50.00                                 & 50.43                                 & 63.46                                 & 40.07                                           \\
                                & SliceGPT                                      & FT2P                                         & 24.58                                 & 26.63                                 & 49.97                                 & 50.00                                 & 25.05                                 & 25.03                                 & 50.00                                 & 24.40                                 & 49.97                                 & 50.00                                 & 32.88                                 & 50.00                                 & 50.00                                 & 50.00                                 & 39.89                                           \\
                                & {\cellcolor[rgb]{1,0.804,0.271}}\textbf{Ours} & {\cellcolor[rgb]{1,0.804,0.271}}\textbf{PAT} & {\cellcolor[rgb]{1,0.804,0.271}}31.19 & {\cellcolor[rgb]{1,0.804,0.271}}38.98 & {\cellcolor[rgb]{1,0.804,0.271}}59.82 & {\cellcolor[rgb]{1,0.804,0.271}}70.00 & {\cellcolor[rgb]{1,0.804,0.271}}32.02 & {\cellcolor[rgb]{1,0.804,0.271}}29.80 & {\cellcolor[rgb]{1,0.804,0.271}}50.74 & {\cellcolor[rgb]{1,0.804,0.271}}47.00 & {\cellcolor[rgb]{1,0.804,0.271}}59.41 & {\cellcolor[rgb]{1,0.804,0.271}}54.15 & {\cellcolor[rgb]{1,0.804,0.271}}36.95 & {\cellcolor[rgb]{1,0.804,0.271}}49.84 & {\cellcolor[rgb]{1,0.804,0.271}}52.57 & {\cellcolor[rgb]{1,0.804,0.271}}22.12 & {\cellcolor[rgb]{1,0.804,0.271}}\textbf{45.33}  \\
\bottomrule
\end{tabular}
}
\end{table*}
\subsection{Detailed HIO Settings}
We present the detailed settings for HIO in \cref{tab:hio-setting}. Our experiments are conducted on the LLaMA 2-7B model, varying the rank values in the LoRA module from 8 to 512 and in the HIO module from 8 to 1024. Across all configurations, our PAT method demonstrates consistently strong performance.
To strike a balance between the number of trainable parameters and computational overhead, we select the ``LoRA20, HIO200'' configuration for the main experiments.
\bibliography{aaai25}
\end{document}